%% file: main.tex
\documentclass{article}
\usepackage{iclr2027_conference,times}

\input{math_commands.tex}

\usepackage{hyperref}
\usepackage{url}
\usepackage{booktabs}
\usepackage{multirow}
\usepackage{colortbl}

\usepackage[table]{xcolor}
\usepackage[most]{tcolorbox}

\usepackage{graphicx}
\usepackage{caption}
\usepackage{booktabs}
\usepackage{tabularx}
\usepackage{array}
\usepackage{wrapfig}
\usepackage{adjustbox}

\definecolor{darkgreen}{RGB}{0,100,0}
\definecolor{darkred}{RGB}{160,30,30}

\newcommand{\gain}[1]{{\scriptsize\textcolor{darkgreen}{$(+#1)$}}}
\newcommand{\drop}[1]{{\scriptsize\textcolor{darkred}{$(-#1)$}}}

\newcommand{\hfmodel}[2]{%
    \href{https://huggingface.co/#1}{#2}%
}

\definecolor{takeawayblue}{RGB}{32,55,85}
\definecolor{takeawaybluebg}{RGB}{244,247,250}
\newtcolorbox[auto counter]{takeaway}[2][]{
    enhanced,
    colback=takeawaybluebg,
    colframe=takeawayblue,
    colbacktitle=takeawayblue,
    coltitle=white,
    title={\textbf{Takeaway~\thetcbcounter:} #2},
    fonttitle=\normalsize,
    fontupper=\normalsize,
    boxrule=0.6pt,
    arc=1.5mm,
    left=6pt,
    right=6pt,
    top=5pt,
    bottom=5pt,
    toptitle=3pt,
    bottomtitle=3pt,
    lefttitle=6pt,
    titlerule=0pt,
    before skip=7pt,
    after skip=7pt,
    #1
}
\title{The Low-Rank Structure of VLA \\ Reinforcement Learning}

\author{
Minjae Oh\thanks{Equal contribution.\hspace{1em}$\dagger$ Corresponding author.},~
Yoonah Park\footnotemark[1],~
Jongwon Lim\footnotemark[1],~
Yohan Jo$^\dagger$ \\
Graduate School of Data Science, Seoul National University \\
\texttt{\{kosair, wisdomsword21, elijah0430, yohan.jo\}@snu.ac.kr}
}

\iclrfinalcopy % Uncomment for camera-ready version, but NOT for submission.
\begin{document}
% Exclude the main text from the table of contents.
\addtocontents{toc}{\protect\setcounter{tocdepth}{-1}}

\maketitle
\lhead{Preprint.}

\begin{abstract}
Reinforcement learning (RL) is increasingly used to post-train vision-language-action (VLA) models, yet how RL reshapes these policies remains poorly understood. We find that RL across widely used flow-based VLA models, including $\pi_{0.5}$ and GR00T~N1.5/N1.6, on LIBERO, ManiSkill, MetaWorld, and CALVIN induces substantially lower-rank parameter updates that are highly concentrated in the action expert's Timestep Modules, a small and previously overlooked component. Through systematic module-replacement experiments, we further show that these modules capture a disproportionate share of the performance gains from RL.
%in $\pi_{0.5}$, keeping only their RL updates yields performance within $1.6\%$ of the full RL policy despite comprising only $27.6\%$ of the action expert's parameters. 
We then characterize what is encoded in these Timestep Modules. First, we show that RL specializes them to the discrete denoising timesteps used during rollouts, and that this discrete-timestep training underlies the low-rank updates. Second, we find that among their outputs, the shift vector changes most distinctly under RL, and through probing, we show that shift update directions strongly predict task success (ROC-AUC up to $99.6\%$). Third, we find that the geometry of shift updates reflects task relationships, as their pairwise similarity correlates with cross-task transfer patterns. 
Building on these findings, we show that steering along shift update directions further improves RL-trained policies without additional RL training. Overall, we provide a systematic understanding of how RL reshapes VLA policies by studying how learned signals are encoded in parameter space, offering insights into more efficient and interpretable VLA post-training.
\end{abstract}

\section{Introduction}
Vision-language-action (VLA) models have emerged as promising general-purpose robot foundation models, demonstrating strong performance across diverse manipulation tasks and embodiments~\citep{zitkovich2023rt,o2024openx,black2024pi0,bjorck2025gr00t}. Similar to other foundation models, VLAs are typically trained in two stages: large-scale pre-training on vision-language and action data, followed by post-training on smaller, task-specific datasets~\citep{kim2025fine,black2024pi0}. As collecting high-quality robot demonstrations is often costly, recent work has demonstrated the effectiveness of reinforcement learning (RL) for VLA post-training in simulation~\citep{tan2025interactive,li2026simplevlarl,chen2025pi_rl,wang2026qwenvla}, building on the success of RL in LLM post-training~\citep{ouyang2022training,guo2025deepseek}. For LLMs, extensive work has characterized the parameter-space structure induced by RL post-training, enabling more effective and efficient learning strategies, including low-rank methods such as LoRA~\citep{schulman2025lora,yuchen2026on,zhang2026geora,yin2026evaluating}. Yet, an analogous understanding of VLA RL remains largely unexplored, particularly given the distinct flow-based action expert architectures commonly used in modern VLAs (\S~\ref{sec:background}). Consequently, it remains unclear where and how RL learning signals are encoded in the parameter space of VLAs.

In this work, we seek to systematically understand how RL learning signals reshape VLAs from a parameter-space perspective. Across widely used flow-based VLA families ($\pi$, GR00T)~\citep{intelligence2025pi_0p5,bjorck2025gr00t} and manipulation benchmarks (LIBERO, ManiSkill, MetaWorld, CALVIN)~\citep{liu2023libero,tao2024maniskill3,yu2020meta,mees2022calvin}, we consistently find that RL induces low-rank parameter updates concentrated in a small, specific module. Surprisingly, the dominant RL updates are concentrated in a previously overlooked component of the action expert, the \emph{Timestep Modules} (Figure~\ref{fig:Timestep_modules}), which condition flow-matching policies on the denoising timestep. Despite exhibiting strongly low-rank updates and thus being natural targets for parameter-efficient adaptation, these modules are typically omitted from standard LoRA configurations. Furthermore, we do not observe the same low-rank structure in flow-based image and video generation models trained with RL, suggesting that this phenomenon is specific to action policies rather than a generic property of flow-based architectures (\S~\ref{sec:parameter_analysis}). Through controlled module-replacement experiments, we show that the Timestep Modules account for a disproportionate share of the performance gains from RL. Consistent with this finding, targeting LoRA specifically to the Timestep Modules outperforms standard LoRA configurations (\S~\ref{sec:Timestep_module_analysis__replacement}).

Building on these findings, we further characterize how the individual components of the Timestep Modules encode RL learning signals. First, unlike standard behavior cloning, RL sharply specializes the Timestep Modules to selectively respond to the discrete denoising timesteps encountered during on-policy learning, and this discrete-timestep training underlies the low-rank updates (\S~\ref{sec:analysis_rl_induced_update}). Second, the learned updates encode task-specific information in a highly compact form due to their low-rank structure, effectively collapsing into a small number of vectors. As a result, one output of the Timestep Modules---the shift vector---carries most of this task-related information (\S~\ref{sec:scale_gate_shift}). We confirm this through probing and task correlation analyses: shift update directions predict downstream task success with an ROC-AUC of up to $99.6\%$, and their similarity patterns correlate strongly with cross-task transfer patterns, with Spearman $\rho=0.794$ (\S~\ref{sec:analysis_shift_direction}).
Finally, since RL induces a distinct low-dimensional update direction for each task, steering RL-trained policies along these directions further improves performance at test time (\S~\ref{sec:exploiting}). Together, we find that the Timestep Modules encode RL learning signals as a small set of timestep-specific vectors that are independent of the image and language inputs or the current task progress, and that their updates alone are sufficient to capture a substantial portion of the performance gains from RL.

Overall, we provide a parameter-level characterization of VLA post-training, showing, to our knowledge for the first time, that RL induces low-rank, structured parameter updates concentrated in the Timestep Modules, driven by training on discrete denoising timesteps. More broadly, our findings lay the groundwork for future work on more parameter-efficient RL post-training, improved adaptation strategies, and the composition and transfer of task-specific RL behaviors.

\section{Background}
\label{sec:background}
\subsection{Flow-based VLA Policies}
\label{sec:background_flow}
Recent VLAs increasingly pair a pretrained vision-language backbone with a dedicated flow-based action expert for continuous robot control~\citep{black2024pi0,intelligence2025pi_0p5,bjorck2025gr00t}. These policies commonly generate \emph{action chunks}, improving temporal consistency~\citep{zhao2023learning,chi2025diffusion} and reducing generation latency compared with discrete action-token policies~\citep{black2024pi0}. Such action experts are trained with a flow-matching objective~\citep{lipman2023flow} to predict the transformation from a noisy action chunk toward the demonstrated action chunk. The standard behavior cloning (BC) flow-matching objective for training the action expert $V_\theta$, parameterized by $\theta$, is:
\begin{equation}
    \mathcal{L}_{\mathrm{FM}}
    =
    \mathbb{E}_{a,\epsilon,\tau}
    \left[
    \left\|
    V_\theta(a_\tau,\tau,o) - (a-\epsilon)
    \right\|_2^2
    \right],
    \qquad
    a_\tau = (1-\tau)\epsilon + \tau a,
\end{equation}
where $a$ is the demonstrated action chunk, $\epsilon$ is sampled noise, and $\tau \sim \mathcal{U}(0,1)$ is the denoising timestep. Because $\tau$ is sampled continuously, the action expert is trained across the full interval $\tau \in [0,1]$. At inference time, the action expert is run at a discrete sequence of timesteps, iteratively transforming an initial noise sample into the final action chunk.

\begin{figure*}[t]
    \centering
    \includegraphics[width=0.7\linewidth]{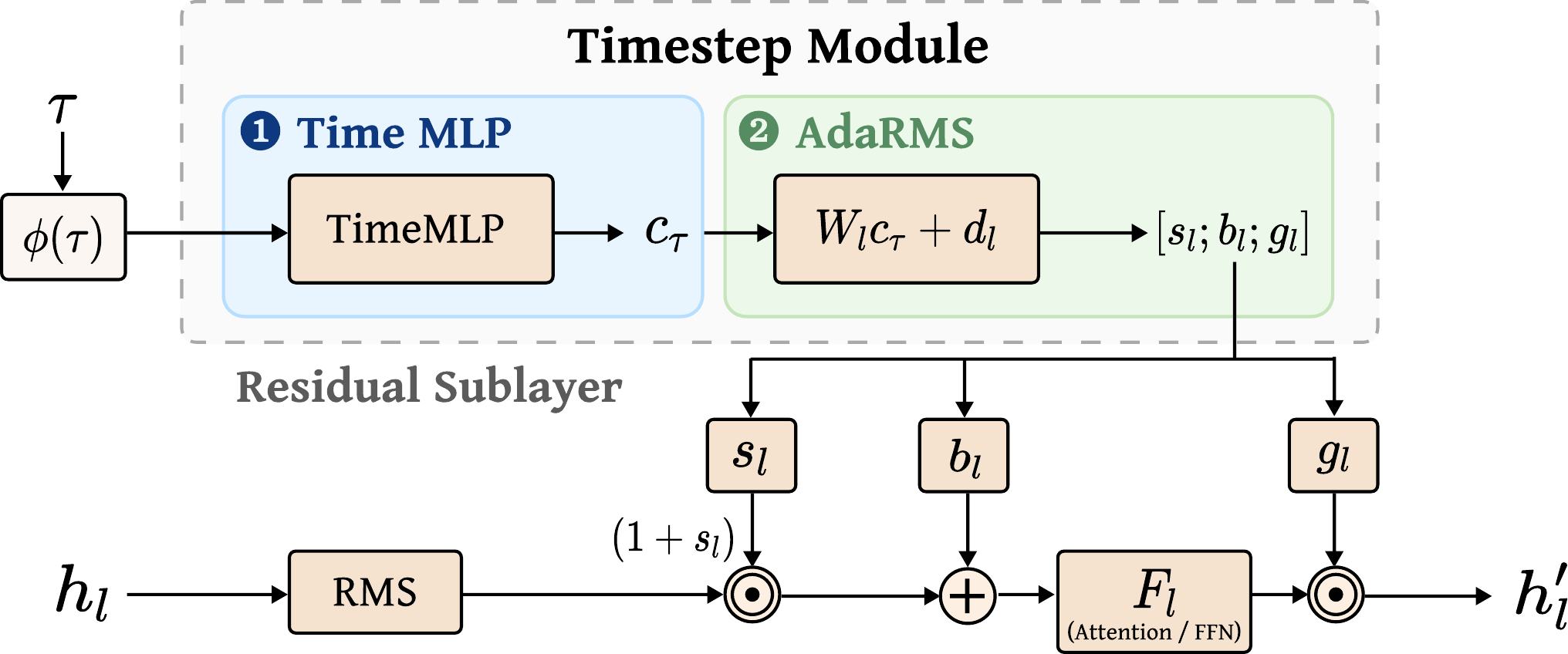}
    \caption{    
    \textbf{Overview of the Timestep Module in \boldmath$\pi_{0.5}$.}
    The timestep embedding is transformed by a TimeMLP and AdaRMS to produce timestep-dependent scale, shift, and gate vectors that modulate each attention or FFN residual sublayer.
    }
    \label{fig:Timestep_modules}
\end{figure*}

\paragraph{Timestep Modules.}
\label{sec:Timestep_modules}
In this work, we use \emph{Timestep Modules} to refer to the components of the action expert that transform the scalar denoising timestep $\tau$ into vector embeddings that modulate the hidden states. For example, a three-step denoising process uses $\tau\in\{0,\frac{1}{3},\frac{2}{3}\}$ as inputs. In the widely used $\pi_{0.5}$ model, the Timestep Modules consist of a Time MLP and adaptive RMS normalization (AdaRMS) (Figure~\ref{fig:Timestep_modules})~\citep{intelligence2025pi_0p5}, which together account for only $27.58\%$ of the action expert's parameters, with most ($98.23\%$) belonging to AdaRMS. During each forward pass, the Time MLP first maps a sinusoidal timestep embedding $\phi(\tau)$ to a conditioning vector $c_\tau$:
\begin{equation}
    c_\tau = \mathrm{TimeMLP}(\phi(\tau)).
\end{equation}
AdaRMS, parameterized by $W_\ell$ and $d_\ell$, then maps $c_\tau$ to scale, shift, and gate vectors $(s_\ell,b_\ell,g_\ell)$:
\begin{align}
    \label{eq:Timestep_modules}
    [s_\ell;b_\ell;g_\ell]
    &= W_\ell c_\tau+d_\ell, \\
    z_\ell
    &= (1+s_\ell)\odot\mathrm{RMS}(h_\ell)+b_\ell, \\
    h'_\ell
    &= h_\ell+g_\ell\odot F_\ell(z_\ell).
\end{align}
Here, the hidden state $h_\ell$ is RMS-normalized, scaled, and shifted to obtain $z_\ell$, which is passed through the corresponding attention or feed-forward sublayer $F_\ell$ and gated before being added back to the residual stream. We collectively refer to the Time MLP and the AdaRMS parameters $\{W_\ell,d_\ell\}$ as the Timestep Modules. Notably, the resulting scale, shift, and gate vectors are conditioned solely on the timestep $\tau$, independent of visual or language inputs. Consequently, with a fixed schedule of $K$ denoising steps, the Timestep Modules produce only $K$ distinct sets of modulation vectors per layer at inference. We further detail the Timestep Modules of various VLAs in \S~\ref{app:vla_architecture}.

\subsection{Reinforcement Learning for VLA}
\label{sec:background_rl}
Recent work has shown that reinforcement learning (RL) can effectively post-train VLAs through online interaction in simulation, reducing reliance on additional expert demonstrations~\citep{li2026simplevlarl,chen2025pi_rl}. Following recent work on flow-based VLA RL~\citep{chen2025pi_rl}, we use proximal policy optimization (PPO)~\citep{schulman2017proximal} as the base RL algorithm:
\begin{equation}
    \mathcal{L}_{\mathrm{PPO}}(\theta)
    =
    -\mathbb{E}_{i}
    \left[
    \min\left(
    \rho_i(\theta)\hat{A}_i,\,
    \operatorname{clip}\left(\rho_i(\theta),1-\varepsilon,1+\varepsilon\right)\hat{A}_i
    \right)
    \right],
\end{equation}
where $\rho_i(\theta)$, $\hat{A}_i$, and $\varepsilon$ are the policy ratio, estimated advantage, and clipping threshold, respectively. Notably, on-policy RL such as PPO performs updates on trajectories generated by the action expert and thus trains the action expert only at the discrete denoising timesteps used during rollouts. This differs from BC, which trains the action expert on continuous timesteps in $[0,1]$ (\S~\ref{sec:background_flow}).

\section{Parameter Updates during VLA Reinforcement Learning}
We begin by examining where and how RL updates the parameters of VLAs. By analyzing the density and effective rank of parameter updates across multiple VLA architectures and benchmarks, we find that (1) RL induces low-rank updates concentrated in the Timestep Modules, an overlooked component of the action expert (\S~\ref{sec:parameter_analysis}), and (2) these low-rank updates account for a disproportionate share of the performance gain from RL, as shown through controlled module-replacement experiments (\S~\ref{sec:Timestep_module_analysis__replacement}).

\begin{figure*}[t]
    \centering
    \includegraphics[width=\textwidth]{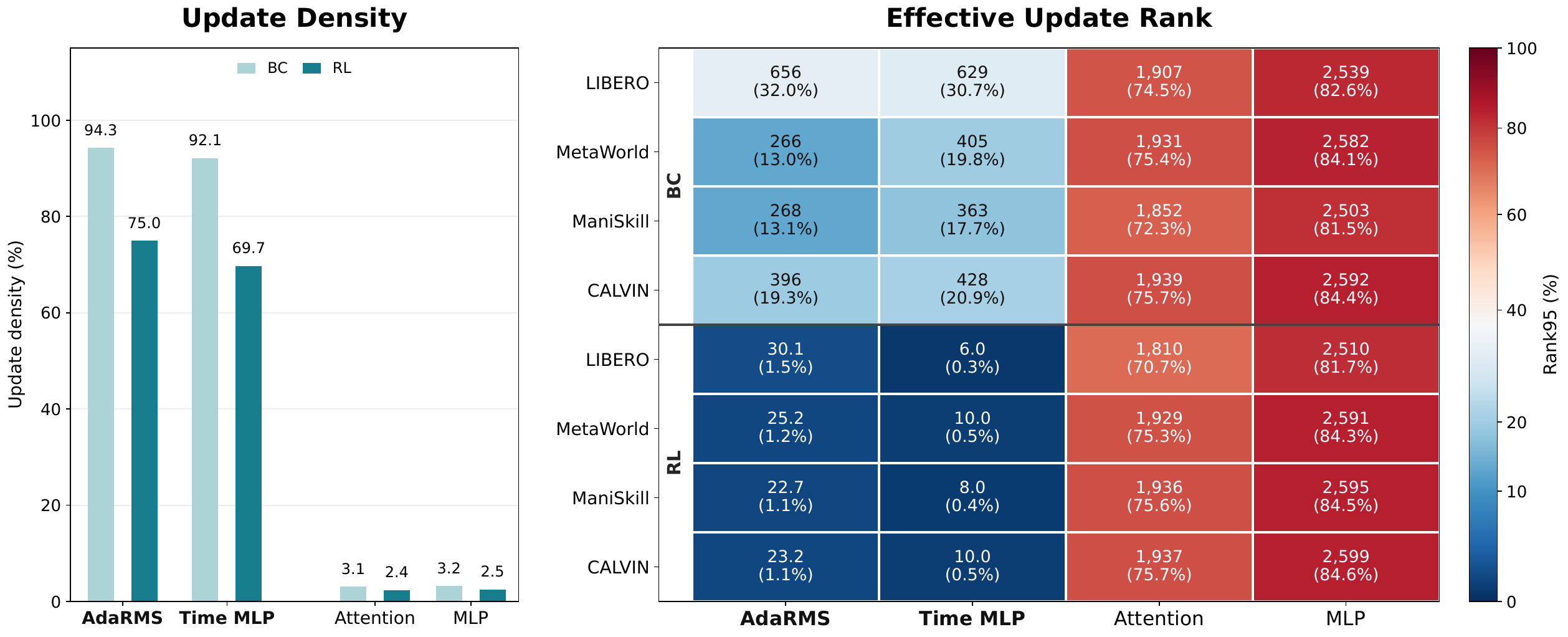}
    \caption{
        \textbf{Timestep Modules receive dense but low-rank RL updates.}
        (Left) Update density for BC and RL checkpoints trained on LIBERO-Spatial from the same $\pi_{0.5}$ policy.
        (Right) Effective update rank across $\pi_{0.5}$ BC and RL checkpoints.
    }
    \label{fig:pi05-update-density-rank}
\end{figure*}

\subsection{Where and How Does RL Update VLA Parameters}
\label{sec:parameter_analysis}  
\paragraph{Settings.} We study both publicly released and in-house-trained BC and RL checkpoints from multiple flow-based VLA families, including $\pi_{0.5}$, GR00T~N1.5, GR00T~N1.6, and, additionally, SmolVLA~\citep{intelligence2025pi_0p5,bjorck2025gr00t, smolvla}, spanning LIBERO, ManiSkill, MetaWorld, and CALVIN~\citep{liu2023libero,tao2024maniskill3,yu2020meta,mees2022calvin} (\S~\ref{app:model_list}).

We define the parameter update as $\Delta W = W_{\mathrm{trained}}-W_{\mathrm{reference}}$, where $W_{\mathrm{trained}}$ and $W_{\mathrm{reference}}$ denote the parameters after and before training, respectively. We define \emph{update density} as the fraction of parameters whose absolute change exceeds a small threshold ($10^{-5}$), following \citet{mukherjee2025reinforcement}, and define the \emph{effective update rank} as the smallest number of singular directions that explain $95\%$ of the update's squared Frobenius norm:
\begin{equation}
    r_{95}(\Delta W)
    =
    \min\left\{
        r:
        \frac{\sum_{i=1}^{r}\sigma_i^2}
             {\sum_i\sigma_i^2}
        \geq 0.95
    \right\},
\end{equation}
where $\sigma_i$ are the singular values of $\Delta W$ in descending order. We further ablate thresholds in \S~\ref{app:threshold_ablations}.

\paragraph{Results.} Figure~\ref{fig:pi05-update-density-rank} (left) shows the update density of each component for BC and RL checkpoints trained on LIBERO-Spatial with $\pi_{0.5}$.
In both cases, the Timestep Modules (AdaRMS and Time MLP) are updated far more densely ($70$--$95\%$) than attention and MLP modules ($<5\%$). Figure~\ref{fig:pi05-update-density-rank} (right) shows the effective rank of the updates across multiple BC and RL checkpoints. Here, BC and RL differ qualitatively, as RL updates to the Timestep Modules have an effective rank of only $6$--$30$, roughly an order of magnitude lower than BC ($250$--$650$), whereas RL updates to attention and MLP modules remain high-rank. We provide additional results in \S~\ref{app:vla_models_parameter_updates}. 

Overall, we find that VLA post-training concentrates its updates in the Timestep Modules, and RL further compresses these updates to a low effective rank. Furthermore, this low-rank structure does not appear in RL-trained image and video DiTs, whose Timestep Module updates require $29$--$80\%$ of the available rank (\S~\ref{app:image_video_parameter_analyses}), suggesting that it is specific to VLA policies rather than a general property of timestep-conditioned models.

Notably, the Timestep Modules produce only the scale, shift, and gate vectors that modulate the hidden states (Eq.~\ref{eq:Timestep_modules}), and these vectors depend solely on the denoising timestep $\tau$. A low-rank update therefore shifts these modulation vectors along only a few fixed directions, regardless of the observation, instruction, or current stage of task execution. This raises the question of whether such simple, input-independent changes can account for the performance gains from RL, which we test next through module replacement.

\begin{takeaway}[label=tk:lowrank]{
    RL learns dense, low-rank updates in VLA Timestep Modules.
}
Across various flow-based VLAs, RL updates the Timestep Modules more densely than other modules and, unlike BC, induces low-rank updates to the Timestep Modules.
\end{takeaway}

\subsection{Do Timestep Modules Capture the RL Gain?}
\label{sec:Timestep_module_analysis__replacement}

\begin{table*}[t] %linenumber
    \centering
    \small
    \renewcommand{\arraystretch}{1.15}

    % Center fixed-width number fields and align their decimal points.
    % The optional argument preserves bold/underline without marking padding.
    \newcommand{\tabledecimal}[3][]{%
        \makebox[18pt][r]{#1{#2}}\makebox[2.5pt][c]{#1{.}}\makebox[11pt][l]{#1{#3}}%
    }
    \setlength{\tabcolsep}{1.7pt}
    \caption{\textbf{Timestep Modules encode more of the RL gain than all other parameters.} Success rates (\%) by condition. LIBERO is averaged over the Spatial, Object, and Goal suites. $^\dagger$GR00T N1.6 is evaluated on LIBERO-Spatial only. Best results are in \textbf{bold}, second-best results are \underline{underlined}, and $\Delta$ rows show the difference from RL.}
    \begin{adjustbox}{max width=\textwidth}
    \begin{tabular*}{\textwidth}{@{\extracolsep{\fill}}lccccccc@{}}
        \toprule
        & \multicolumn{4}{c}{\textbf{$\boldsymbol{\pi}_{0.5}$}}
        & \multicolumn{1}{c}{{\footnotesize\shortstack{\textbf{GR00T} \textbf{N1.5}}}}
        & \multicolumn{1}{c}{{\footnotesize\shortstack{\textbf{GR00T} \textbf{N1.6}}}} & \\
        \cmidrule(lr){2-5} \cmidrule(lr){6-6} \cmidrule(lr){7-7}
        \textbf{Condition} & \multicolumn{1}{c}{\textbf{LIBERO}} & \multicolumn{1}{c}{\textbf{ManiSkill}} & \multicolumn{1}{c}{\textbf{MetaWorld}} & \multicolumn{1}{c}{\textbf{CALVIN}} & \multicolumn{1}{c}{\textbf{LIBERO}} & \multicolumn{1}{c}{\textbf{LIBERO}$^\dagger$} & \multicolumn{1}{c@{}}{\textbf{Avg.}} \\
        \midrule

        Base
            & \normalsize \tabledecimal{87}{7} & \normalsize \tabledecimal{42}{2} & \normalsize \tabledecimal{42}{8} & \normalsize \tabledecimal{61}{8} & \normalsize \tabledecimal{54}{4} & \normalsize \tabledecimal{75}{6} & \normalsize \tabledecimal{60}{8} \\
        \midrule
        RL
            & \normalsize \tabledecimal[\textbf]{96}{5} & \normalsize \tabledecimal[\textbf]{89}{1} & \normalsize \tabledecimal[\underline]{69}{2} & \normalsize \tabledecimal[\textbf]{87}{7} & \normalsize \tabledecimal[\textbf]{90}{3} & \normalsize \tabledecimal[\textbf]{85}{0} & \normalsize \tabledecimal[\textbf]{86}{3} \\
        \hspace{0.5em}\textit{Timestep Modules only}
            & \normalsize \tabledecimal[\underline]{95}{1} & \normalsize \tabledecimal[\underline]{87}{5} & \normalsize \tabledecimal[\textbf]{69}{6} & \normalsize \tabledecimal[\underline]{87}{0} & \normalsize \tabledecimal[\underline]{61}{3} & \normalsize \tabledecimal[\underline]{84}{8} & \normalsize \tabledecimal[\underline]{80}{9} \\
        \hspace{1.5em}$\Delta$ vs.\ RL
            & \scriptsize \textcolor{darkred}{\tabledecimal{($-$1}{4)}} & \scriptsize \textcolor{darkred}{\tabledecimal{($-$1}{6)}} & \scriptsize \textcolor{darkgreen}{\tabledecimal{($+$0}{4)}} & \scriptsize \textcolor{darkred}{\tabledecimal{($-$0}{7)}} & \scriptsize \textcolor{darkred}{\tabledecimal{($-$29}{0)}} & \scriptsize \textcolor{darkred}{\tabledecimal{($-$0}{2)}} & \scriptsize \textcolor{darkred}{\tabledecimal{($-$5}{4)}} \\
        \hspace{0.5em}\textit{MLP+Attn only}
            & \normalsize \tabledecimal{90}{5} & \normalsize \tabledecimal{49}{4} & \normalsize \tabledecimal{54}{6} & \normalsize \tabledecimal{62}{7} & \normalsize \tabledecimal{57}{9} & \normalsize \tabledecimal{84}{6} & \normalsize \tabledecimal{66}{6} \\
        \hspace{1.5em}$\Delta$ vs.\ RL
            & \scriptsize \textcolor{darkred}{\tabledecimal{($-$6}{0)}} & \scriptsize \textcolor{darkred}{\tabledecimal{($-$39}{7)}} & \scriptsize \textcolor{darkred}{\tabledecimal{($-$14}{6)}} & \scriptsize \textcolor{darkred}{\tabledecimal{($-$25}{0)}} & \scriptsize \textcolor{darkred}{\tabledecimal{($-$32}{4)}} & \scriptsize \textcolor{darkred}{\tabledecimal{($-$0}{4)}} & \scriptsize \textcolor{darkred}{\tabledecimal{($-$19}{7)}} \\
        \midrule
        Timestep Modules Size
            & \multicolumn{4}{c}{\normalsize 27.58\%} & \normalsize 14.83\% & \normalsize 16.50\% & --- \\
        \bottomrule
    \end{tabular*}
    \end{adjustbox}
    \label{tab:module_replacement}
\end{table*}

\paragraph{Settings.} Building on Takeaway~\ref{tk:lowrank}, we test whether the Timestep Modules actually account for the performance gains from RL. Using the RL-trained $\pi_{0.5}$, GR00T~N1.5, and GR00T~N1.6 checkpoints from \S~\ref{sec:parameter_analysis}, we perform module-replacement experiments. Starting from each RL checkpoint, we either (1) replace the attention and FFN parameters with their base parameters, keeping only the Timestep Modules RL-trained (\textit{Timestep Modules only}), or (2) replace the Timestep Modules with their base parameters, keeping all other RL-trained parameters fixed (\textit{MLP+Attn only}).

\paragraph{Results.} As shown in Table~\ref{tab:module_replacement}, \textit{Timestep Modules only} consistently preserves more of the RL gain than \textit{MLP+Attn only}, despite the Timestep Modules comprising only 14.83--27.58\% (\S~\ref{app:vla_architecture}) of the parameters. Notably, in $\pi_{0.5}$, \textit{Timestep Modules only} nearly matches the full RL policy, reaching $69.6\%$ on MetaWorld and falling behind by only $1.4\%$, $1.6\%$, and $0.7\%$ on LIBERO, ManiSkill, and CALVIN, respectively, whereas \textit{MLP+Attn only} drops by up to $39.7\%$ on ManiSkill.

We further verify this with two additional experiments. First, under the \textit{Timestep Modules only} setting, reconstructing the Timestep Module updates from only their top four singular directions retains most of the performance, while removing these directions largely degrades it, highlighting that the low-rank updates carry most of the RL gain (\S~\ref{app:rank_ablation_analyses}). Second, we compare LoRA on the Timestep Modules against LoRA on all other modules, the standard setting, and find that Timestep-targeted LoRA converges faster and to a higher success rate (\S~\ref{app:lora_experiment}).

Overall, these results indicate that a large portion of the performance gain from RL is captured by low-rank updates to the Timestep Modules. Since these modules only produce the scale, shift, and gate vectors, which depend solely on the denoising timestep and not on the image, instruction, or task progress, this suggests that much of what RL learns is a simple, timestep-specific change to how the action expert is modulated at each denoising step.

\begin{takeaway}[label=tk:module_replacement]{
    Timestep Modules encode a disproportionately large share of the RL gain.
}
Through controlled module-replacement experiments, we show that RL encodes much of its gain in the Timestep Modules, largely through low-rank updates.
\end{takeaway}

\section{What Do the Timestep Modules Encode?}
\label{sec:timestep_module_analysis__details}
In Takeaways~\ref{tk:lowrank} and~\ref{tk:module_replacement}, we showed that RL updates to the Timestep Modules exhibit a low-rank structure that captures a substantial portion of the RL gain. In this section, we examine AdaRMS, the main component of the Timestep Modules: (1) how it specializes to the denoising timesteps under RL, which underlies the low-rank updates (\S~\ref{sec:analysis_rl_induced_update}), and (2) how the scale, shift, and gate vectors change under RL, with the shift vectors standing out (\S~\ref{sec:scale_gate_shift}).

\begin{figure}[t]
    \centering
    \includegraphics[width=\linewidth]
    {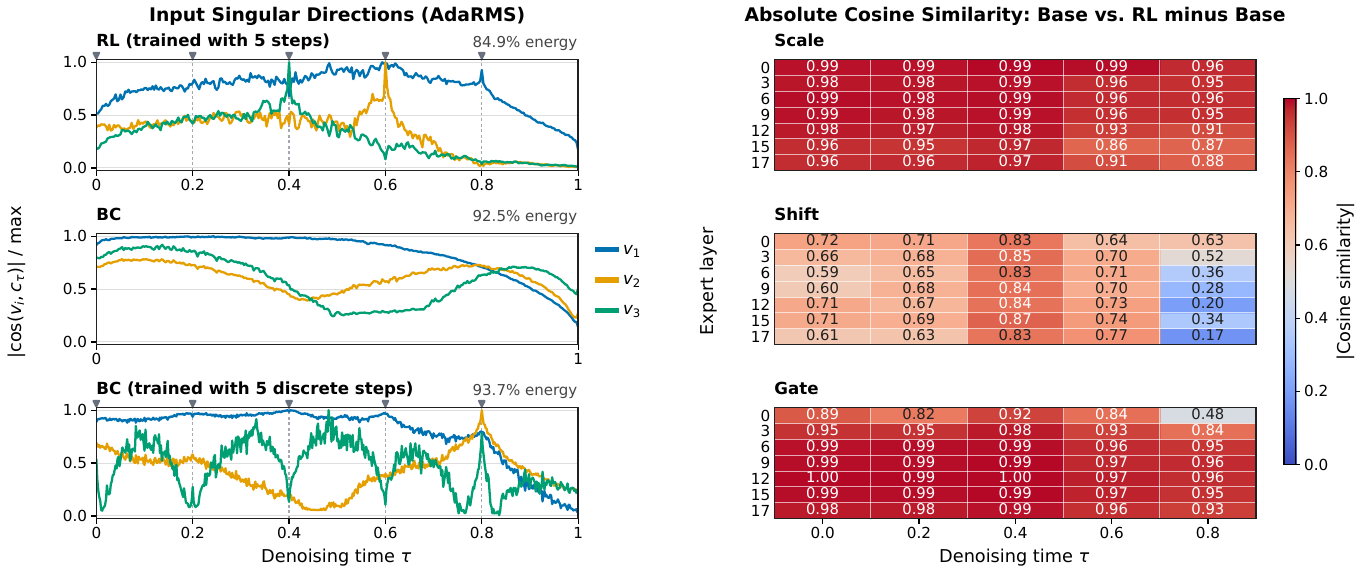}
    \caption{
        \textbf{AdaRMS updates specialize to denoising timesteps, largely changing the shift vector.} (Left) Top three input singular directions of LIBERO-Spatial AdaRMS updates for BC and RL. Unlike standard BC, RL and discrete-timestep BC checkpoints peak near the denoising timesteps used during training. (Right) Absolute cosine similarity between the base AdaRMS output and the RL-induced change (RL $-$ base) in MetaWorld for the scale, shift, and gate vectors, respectively.
    }
    \label{fig:adarms-timestep-analysis}
\end{figure}

\subsection{Specialization to Discrete Timesteps}
\label{sec:analysis_rl_induced_update}
\paragraph{Settings.} As discussed in \S~\ref{sec:background_flow}, the Timestep Modules take only the denoising timestep as input, without any image, text, or task progress. We therefore analyze how AdaRMS learns to respond across denoising timesteps under standard BC and RL, and additionally under BC trained only on the discrete timesteps used during RL (\emph{discrete-timestep BC}) for comparison. Specifically, we compute the SVD of each AdaRMS update $\Delta W_\ell$ to extract the top three input singular directions $v_i$, which account for most of the update energy. We then measure the normalized absolute cosine similarity between each $v_i$ and the conditioning vector $c_\tau$ across denoising timesteps to examine how the AdaRMS update responds to each timestep.

\paragraph{Results.} Figure~\ref{fig:adarms-timestep-analysis} (left) shows the results for $\pi_{0.5}$ on LIBERO-Spatial under standard BC, RL, and discrete-timestep BC. Under standard BC, the similarity varies smoothly across timesteps, whereas RL produces sharply localized responses around the discrete timesteps used during training. Interestingly, this matches the difference in training discussed in \S~\ref{sec:background_rl}, as BC trains on all timesteps $\tau \in [0,1]$, while on-policy RL trains only on the fixed timesteps used during rollouts. Furthermore, discrete-timestep BC exhibits similarly localized responses, and its updates are likewise low-rank (\S~\ref{app:vla_models_parameter_updates}), indicating that discrete-timestep training is the main cause of the low-rank structure. We provide additional results in \S~\ref{app:time_conditioning}. 

This suggests that RL learns timestep-specific modulation patterns for the discrete timesteps used during rollouts. Since the scale, shift, and gate vectors produced by the Timestep Modules capture much of the RL gain (Takeaway~\ref{tk:module_replacement}), the AdaRMS update only needs to act on a few distinct timesteps, for which a low-rank update suffices.

\begin{takeaway}[label=tk:specialization]{RL specializes AdaRMS to the discrete denoising timesteps.}
Unlike standard BC, RL trains AdaRMS only at the few timesteps used during rollouts, causing the module to specialize to those timesteps and resulting in low-rank updates.
\end{takeaway}

\subsection{Disentangling Scale, Shift, and Gate}
\label{sec:scale_gate_shift}
\paragraph{Settings.} We examine how the three AdaRMS output vectors (scale, shift, and gate) change due to RL. We compute the absolute cosine similarity between each output's base value and its RL-induced change (RL $-$ base) at each denoising timestep used during RL training. Values near $1$ indicate rescaling along the existing direction, while values near $0$ indicate a new direction.

\paragraph{Results.} As shown in Figure~\ref{fig:adarms-timestep-analysis} (right), the RL-induced change is substantially more aligned with the base output for scale and gate vectors than for shift vectors, and varies more across denoising timesteps than across Transformer sublayers. This suggests that RL largely rescales the existing scale and gate patterns while introducing new directions predominantly in the shift vectors, which may therefore encode more information learned through RL.

\section{Understanding the Shift Vector}
We find that RL induces low-rank updates concentrated in the Timestep Modules, with the shift vector changing most among their outputs. In this section, we examine what information these shift updates capture (\S~\ref{sec:analysis_shift_direction}) and whether they can be exploited to steer task outcomes (\S~\ref{sec:exploiting}).

\subsection{What Do the Shift Updates Encode?}
\label{sec:analysis_shift_direction}

\subsubsection{Do Shift Updates Encode Task Outcomes?}
\label{sec:probing}

\begin{table}[t]
    \centering
    \small
    \setlength{\tabcolsep}{4.5pt}
    \renewcommand{\arraystretch}{0.95}
    \caption{\textbf{Linear probing of shift vectors in \boldmath$\pi_{0.5}$.}
    F1 and ROC-AUC (\%) for probes applied to the base and RL policies.
    Random denotes probes trained with shuffled outcome labels.}
    \label{tab:shift-probe}

    \begin{tabular}{@{}llcccc@{}}
        \toprule
        & &
        \multicolumn{2}{c}{\textbf{Base Policy}} &
        \multicolumn{2}{c}{\textbf{RL Policy}} \\
        \cmidrule(lr){3-4} \cmidrule(lr){5-6}
        Benchmark & Condition
        & F1 & AUC
        & F1 & AUC \\
        \midrule

        \multirow{2}{*}{LIBERO-Spatial}
        & \textbf{Ours}
        & \textbf{97.1} & \textbf{99.6}
        & \textbf{96.3} & \textbf{96.6} \\
        & Random
        & $58.0 \pm 20.5$ & $46.5 \pm 4.0$
        & $75.2 \pm 13.6$ & $51.5 \pm 3.9$ \\
        \midrule

        \multirow{2}{*}{LIBERO-Object}
        & \textbf{Ours}
        & \textbf{98.2} & \textbf{98.6}
        & \textbf{97.3} & \textbf{97.4} \\
        & Random
        & $76.3 \pm 10.2$ & $53.4 \pm 6.3$
        & $86.2 \pm 15.8$ & $47.2 \pm 13.3$ \\
        \midrule

        \multirow{2}{*}{LIBERO-Goal}
        & \textbf{Ours}
        & \textbf{77.5} & \textbf{98.9}
        & \textbf{74.9} & \textbf{68.0} \\
        & Random
        & $47.6 \pm 37.2$ & $49.0 \pm 3.5$
        & $65.4 \pm 28.9$ & $52.6 \pm 6.4$ \\
        \midrule

        \multirow{2}{*}{ManiSkill}
        & \textbf{Ours}
        & \textbf{95.8} & \textbf{99.6}
        & \textbf{98.3} & \textbf{98.3} \\
        & Random
        & $42.9 \pm 8.9$ & $51.8 \pm 7.0$
        & $87.1 \pm 3.4$ & $46.5 \pm 9.3$ \\
        \midrule

        \multirow{2}{*}{MetaWorld}
        & \textbf{Ours}
        & \textbf{58.3} & \textbf{74.2}
        & \textbf{81.1} & \textbf{90.1} \\
        & Random
        & $43.4 \pm 7.8$ & $49.4 \pm 4.8$
        & $64.8 \pm 6.6$ & $49.4 \pm 6.1$ \\

        \bottomrule
    \end{tabular}
\end{table}

\paragraph{Settings.} 
Because RL improves task success partly through shift updates that are directly added to the hidden state, we hypothesize that the shift update directions capture success-relevant information in the representation space. 
This connects naturally to prior probing work in LLMs, which projects hidden representations onto specific directions to reveal answer correctness \citep{zhang2025reasoning,cencerrado2026no}, as well as recent VLA work on task success prediction \citep{gu2026safe,zhang2026frozen}. 
Following this perspective, we use each RL-induced shift update, $\Delta b = b_{\mathrm{RL}} - b_{\mathrm{base}}$, as a probing direction in representation space and test whether hidden-state projections onto this direction predict the episode's eventual success or failure.

For each checkpoint, we first compute a separate shift update for every attention or MLP sublayer--timestep pair. 
During rollouts, we extract the hidden representation at each corresponding sublayer input, project it onto $\Delta \mathbf{b}$, and average the projection scores over the episode. The resulting episode-level scores across all sublayer--timestep pairs form a feature vector, which an $\ell_2$-regularized logistic regression probe combines across sublayers and timesteps to predict episode success or failure.

We evaluate both the base and RL $\pi_{0.5}$ policies on LIBERO-Spatial, LIBERO-Object, LIBERO-Goal, ManiSkill, and MetaWorld. Within each benchmark, tasks are split 70:30 into training and test sets, and the probe is trained on the training tasks and evaluated on held-out tasks.

\paragraph{Results.} As shown in Table~\ref{tab:shift-probe}, shift update projections strongly predict episode outcomes, reaching ROC-AUCs of $98.6$--$99.6$ across LIBERO-Spatial, LIBERO-Object, LIBERO-Goal, and ManiSkill, well above random-label controls. Notably, our probes are predictive for both the base and RL policies, suggesting that RL aligns with success-relevant structure already present in the representation space and preserves this alignment after optimization. Even a single sublayer--timestep direction yields ROC-AUCs of $96.9$--$99.0$ on LIBERO-Spatial, LIBERO-Object, and ManiSkill. Although the sign of the association varies across sublayer--timestep pairs, predictive performance remains consistently strong (\S~\ref{app:probing_results}). Overall, these results suggest that RL-induced shift updates are structured around outcome-relevant directions in the representation space.

\subsubsection{Does Shift Update Geometry Encode Task Relationships?}

\begin{wrapfigure}[30]{r}{0.42\textwidth}
    \centering
    \includegraphics[width=0.88\linewidth]{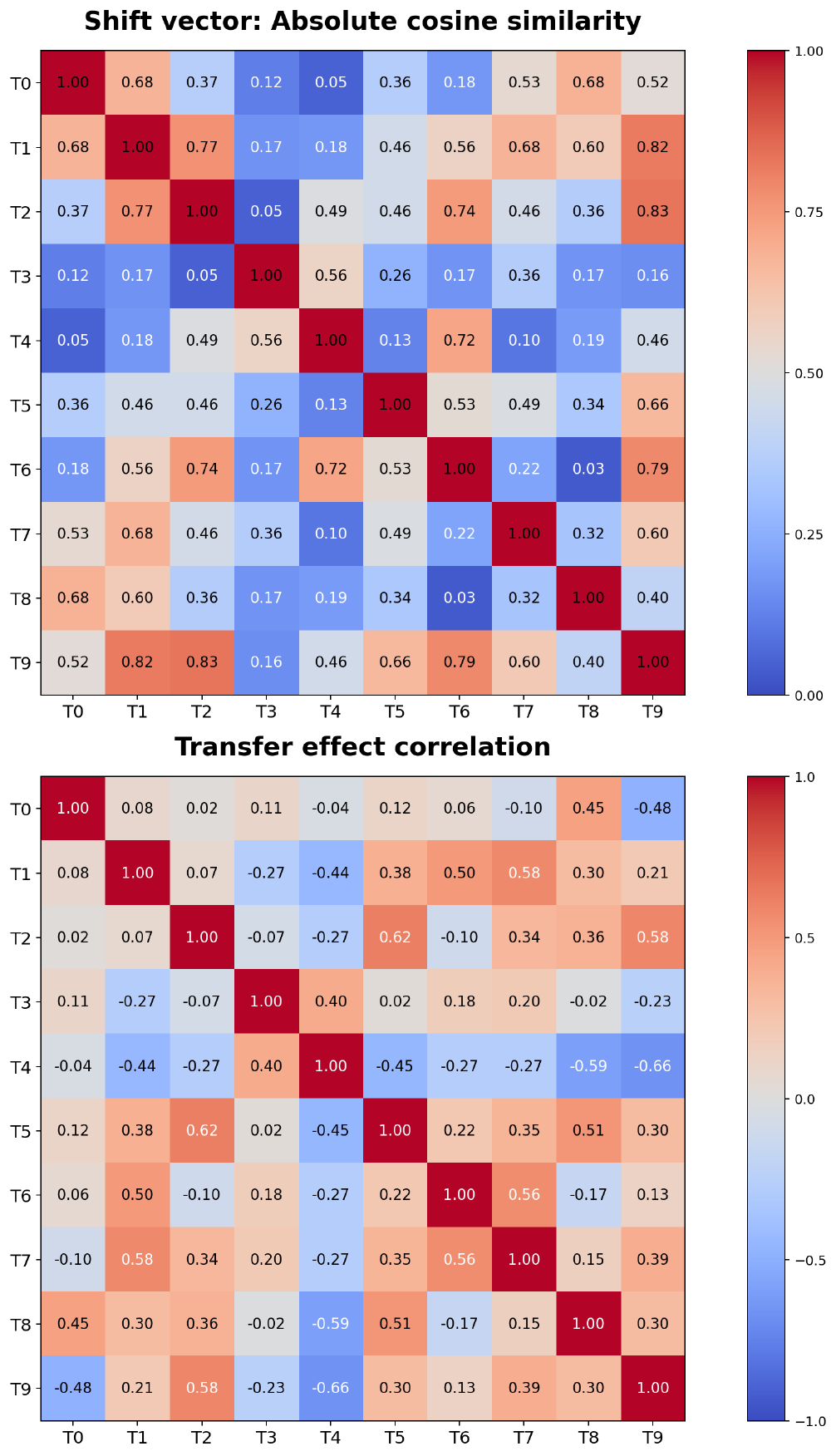}
    \caption{\textbf{Cross-task structure in \boldmath$\pi_{0.5}$ on LIBERO-Spatial subtasks.} (Top) Absolute cosine similarity between shift vectors. (Bottom) Correlation of cross-task transfer effects.}
    \label{fig:10taskvec}
\end{wrapfigure}

\paragraph{Settings.} We next ask whether shift update geometry reflects relationships across tasks. We hypothesize that related tasks exhibit aligned shift updates and similar cross-task transfer behavior. We train a separate RL policy for each of the ten LIBERO-Spatial tasks and construct two vectors for each task: a \emph{shift-update vector} from its single-task RL policy and a \emph{transfer-effect vector} containing the success-rate gains of all ten policies over the base policy on that task. We then compare tasks using each of these two vectors. For shift-update vectors, we measure pairwise alignment using absolute cosine similarity, since vectors pointing in opposite directions can still encode similar task-related information. For transfer-effect vectors, we measure pairwise similarity using Spearman correlation. Finally, for each task, we compute the Spearman correlation between its shift-update-vector similarities and transfer-effect-vector similarities to the other tasks. This tests whether tasks with geometrically aligned shift updates also exhibit similar cross-task transfer behavior.

\paragraph{Results.} We find that shift-update alignment is positively associated with cross-task transfer behavior, with a mean task-wise correlation of $0.504$ (range: $0.224$--$0.794$). Figure~\ref{fig:10taskvec} further shows that the pairwise structure of shift-update alignment (top) closely resembles that of cross-task transfer similarity (bottom). These results suggest that the geometry of shift update directions reflects relationships among the tasks on which they are learned, providing a compact representation of how RL updates may transfer across tasks.

\subsection{Exploiting the Shift Updates}
\label{sec:exploiting}

\begin{wraptable}[12]{r}{0.5\textwidth}
        \centering
        \small
        \renewcommand{\arraystretch}{0.9}
        \setlength{\tabcolsep}{5pt}
    \caption{\textbf{{\boldmath $\pi_{0.5}$} adaptive steering results.} Success rates (\%) on the 30\% test split for RL, random vector steering, and shift vector steering.}
    \begin{tabular}{lccc}
    \toprule
    Checkpoint & RL & Random & Steered \\
    \midrule
    LIBERO-Spatial
    & 86.7
    & $87.2 \pm 0.3$
    & \textbf{90.0} \gain{3.3} \\
    
    LIBERO-Object
    & 94.7
    & $94.7 \pm 0.0$
    & \textbf{96.7} \gain{2.0} \\
    
    LIBERO-Goal
    & 90.7
    & $90.7 \pm 0.0$
    & \textbf{94.7} \gain{4.0} \\
    
    ManiSkill
    & 90.0
    & $88.4 \pm 3.4$
    & \textbf{92.0} \gain{2.0} \\
    
    MetaWorld
    & 66.7
    & $64.1 \pm 1.8$
    & \textbf{68.7} \gain{2.0} \\
    \bottomrule
    \end{tabular}
    \label{tab:shift-steering}
\end{wraptable}

\paragraph{Settings.} Having shown that shift updates encode task outcomes, we ask whether this information can be exploited to improve policy performance. Specifically, we steer representations along the shift update direction toward values associated with successful episodes. We select the sublayer--timestep pair with the highest probing F1 score and perform steering along its shift update direction. Inspired by prior adaptive activation steering methods for LLMs \citep{cheng2026liseco,park2026bridging}, we intervene only when the probe indicates that the projection deviates from the mean projection of successful episodes, shifting it by the minimum amount needed to reach this mean. We use the same models, benchmarks, and split as in \S~\ref{sec:probing}.

\paragraph{Results.} As shown in Table~\ref{tab:shift-steering}, adaptive steering improves the final RL policy across all five benchmarks by $2.0$--$4.0\%$ and consistently outperforms matched-norm random directions. For example, success rates increase from $86.7\%$ to $90.0\%$ on LIBERO-Spatial and from $90.7\%$ to $94.7\%$ on LIBERO-Goal. Similar gains from steering the base policy further support the causal role of the learned shift directions (\S~\ref{app:steering_results}). Overall, these results suggest that the shift updates capture an actionable control signal that is not fully exploited by the final RL policy and can be further leveraged at inference time to improve an already trained policy.

\begin{takeaway}{
Shift updates largely encode task-relevant RL signals.
}
Shift updates encode task outcomes and cross-task relationships, enabling test-time steering to further improve the final RL policy.
\end{takeaway}

\section{Related Work}
\subsection{Vision-Language-Action Models}
Vision-language-action models build on pretrained vision-language models to map visual observations and language instructions to robot actions~\citep{zitkovich2023rt,o2024openx,kim2025fine}. Recent VLAs increasingly adopt dedicated diffusion- or flow-based action experts~\citep{black2024pi0,bjorck2025gr00t,intelligence2025pi_0p5}, which, combined with action chunking, generate sequences of actions jointly, improving temporal coherence and inference latency~\citep{zhao2023learning}. 
In this work, we focus on flow-based VLAs and investigate how RL post-training reshapes their action experts, particularly the Timestep Modules.
Our work is closely related to recent efforts to understand VLA internal representations through probing and representation steering~\citep{haon2025mechanistic,wang2026vla,gu2026safe,buurmeijer2026observing}, but differs in focusing on how RL reshapes these representations.

\subsection{Reinforcement Learning}
Inspired by the success of RL in LLMs~\citep{ouyang2022training,guo2025deepseek}, recent work has successfully post-trained VLAs through online RL in simulation~\citep{tan2025interactive, li2026simplevlarl, liu2025what}. In the LLM RL literature, rigorous studies have examined RL at the parameter level, identifying sparse parameter updates~\citep{mukherjee2025reinforcement} and, in other settings, dominant low-rank update directions~\citep{yuchen2026on}. Other work shows that RL can succeed by updating only a few parameters, motivating more parameter-efficient post-training~\citep{morris2026learning,yin2026evaluating,schulman2025lora,zhang2026geora}. While recent work has proposed various recipes for optimizing flow-based policies~\citep{liu2026flow, mcallister2026flow}, including the action experts of recent VLAs~\citep{chen2025pi_rl}, what exactly happens to the action expert during RL has not yet been studied, which we address in this work.

\section{Conclusion and Discussion}
In this work, we studied how RL reshapes flow-based VLAs from a parameter-space perspective across $\pi_{0.5}$ and GR00T on LIBERO, ManiSkill, MetaWorld, and CALVIN. We consistently find, to our knowledge for the first time, that RL encodes its learning signal in low-rank updates to the Timestep Modules, a small and previously overlooked component of the action expert. Leveraging this structure, we further show through probing and steering experiments that much of the RL signal is captured at the vector level and concentrated in one of their outputs, the shift vector. Together, our results suggest that much of RL post-training in VLAs can be explained by simple, timestep-specific vector-level modulations of the action expert's hidden states.

Our findings primarily motivate more parameter-efficient RL post-training and suggest several directions for future work. First, since discrete-timestep BC induces a similar low-rank structure (\S~\ref{sec:analysis_rl_induced_update}), investigating its effects and potential advantages over standard continuous-timestep BC is a promising direction for VLA policies. Second, our results may help explain the effectiveness of residual RL~\citep{johannink2019residual} and related approaches that adapt frozen VLAs through lightweight trainable modules~\citep{xiao2026pld,wagenmaker2025steering,xu2026rlt}. If full-parameter RL largely reduces to low-rank modulation, a small residual module may suffice to capture much of its benefit, potentially informing future residual RL designs.

\subsection*{AI Use Statement}
In this work, we used generative AI assistants to edit the manuscript for clarity and grammar, format tables and \LaTeX{} code, prototype figures, and assist with parts of our analysis and plotting code. We did not use generative AI tools to propose or refine hypotheses, design or provide feedback on research methods or experiments, or interpret results. Generating synthetic datasets, developing theoretical models or conceptual frameworks, formulating mathematical claims, proving mathematical claims or assisting with the writing of proofs, translation, dataset cleaning or reformatting, and qualitative or thematic data analysis are not applicable to this work. Additionally, we used generative AI tools to propose a title or keywords, identify relevant literature, and source or search for information. The authors reviewed and rewrote all AI-assisted text and reviewed all AI-assisted code for correctness. We take responsibility for the final content of this work, including text, claims, and artifacts produced with the aid of generative AI.

\subsection*{Reproducibility statement}
We plan to publicly release the code used in our experiments, along with relevant artifacts such as our trained checkpoints, upon acceptance. To support future reproduction, we provide the detailed hyperparameters and settings used throughout our experiments in \S~\ref{app:exp_settings}.



\bibliography{iclr2027_conference}
\bibliographystyle{iclr2027_conference}

\clearpage
\appendix

% Include appendix sections and subsections.
\addtocontents{toc}{\protect\setcounter{tocdepth}{2}}

\begingroup
\renewcommand{\contentsname}{Appendix Contents}
\tableofcontents
\endgroup

\clearpage
\section{VLA Architectures and Timestep Modules}
\label{app:vla_architecture}
In this section, we detail the Timestep Modules in the action experts of the flow-based VLAs $\pi_{0.5}$, GR00T N1.5/N1.6, and SmolVLA. Table~\ref{tab:timestep-architecture} summarizes each model's architecture and parameter counts.

\begin{table}[!htb]
    \centering
    \small
    \setlength{\tabcolsep}{4pt}
    \renewcommand{\arraystretch}{1.15}
    \caption{
    \textbf{Architecture and size of Timestep Modules.} For GR00T, the count includes the shared Time MLP, the adaptive normalization projections, and the timestep-input blocks of the embodiment-specific action encoders. For SmolVLA, it includes only the timestep-input block of the action--timestep MLP, not the full joint MLP.
    }
    \label{tab:timestep-architecture}
    \begin{tabular*}{\linewidth}{@{\extracolsep{\fill}}lcccc@{}}
        \toprule
        & $\pi_{0.5}$ & GR00T N1.5 & GR00T N1.6 & SmolVLA \\
        \midrule
        Expert hidden size          & 1,024 & 1,536 & 1,536 & 720 \\
        Adaptive normalization      & AdaRMSNorm & AdaLayerNorm & AdaLayerNorm & -- \\
        Modulation outputs          & scale, shift, gate & scale, shift & scale, shift & -- \\
        \midrule
        Timestep parameters (M)     & 118.60 & 158.52 & 234.07 & 0.52 \\
        Action expert parameters (M)  & 430.10 & 1,068.81 & 1,418.62 & 99.88 \\
        Timestep share of action expert (\%) & 27.58 & 14.83 & 16.50 & 0.52 \\
        \bottomrule
    \end{tabular*}
\end{table}

\paragraph{Scope and counting convention.}
We define Timestep Modules as the parameters that directly map timestep embeddings to conditioning vectors, such as AdaRMS projections, and thus exclude parameters that do not transform the timestep embeddings, such as attention and feed-forward layers. For modules that take both timestep and action embeddings, we count only the weights corresponding to the timestep input and exclude the action-input block, the joint bias, and downstream projections of mixed features. We count shared parameters once, even if they are reused during inference.

\subsection{\texorpdfstring{$\pi_{0.5}$}{pi0.5}: Time MLP and AdaRMS}
The Timestep Modules of $\pi_{0.5}$~\citep{intelligence2025pi_0p5} consist of a shared Time MLP and layer-specific adaptive RMS normalization (AdaRMS) projections. The action expert has $18$ transformer layers with hidden size $d=1024$. The Time MLP maps a sinusoidal timestep embedding to a conditioning vector shared across layers:
\begin{equation}
    c_\tau
    =
    \mathrm{SiLU}\!\left(
        W_2\,\mathrm{SiLU}(W_1\phi(\tau)+b_1)+b_2
    \right),
\end{equation}
where $W_1,W_2\in\mathbb{R}^{1024\times1024}$.

The AdaRMS modules then produce three conditioning vectors: the scale, shift, and gate vectors. Notably, each transformer layer contains two different AdaRMS modules, one before the attention sublayer and the other before the feed-forward sublayer. For sublayer $\ell$, with attention or feed-forward transformation $F_\ell$, AdaRMS operates as follows:
\begin{align}
    [s_\ell;b_\ell;g_\ell]
        &= W_\ell c_\tau+d_\ell, \\
    z_\ell
        &= (1+s_\ell)\odot\mathrm{RMS}(h_\ell)+b_\ell, \\
    h'_\ell
        &= h_\ell+g_\ell\odot F_\ell(z_\ell).
\end{align}

Each AdaRMS projection has shape $3072\times1024$, consisting of equal-size $1024\times1024$ blocks for its three outputs. As described in Table~\ref{tab:timestep-architecture}, Timestep Modules account for $27.58\%$ of the action expert, with AdaRMS accounting for the majority ($98.23\%$).

\subsection{GR00T: AdaLayerNorm and Action--Timestep Encoding}
The Timestep Modules of GR00T N1.5 and N1.6 incorporate timestep information through two mechanisms: adaptive LayerNorm and a joint action--timestep encoder. The action experts of GR00T N1.5 and N1.6 have $16$ and $32$ transformer layers, respectively, both with hidden size $d=1536$.

\paragraph{Time MLP and adaptive LayerNorm.} Similar to $\pi_{0.5}$, a shared Time MLP maps a $256$-dimensional sinusoidal timestep embedding to a $1536$-dimensional conditioning vector:
\begin{equation}
    c_\tau
    =
    W_2\,\mathrm{SiLU}(W_1\phi(\tau)+b_1)+b_2,
\end{equation}
where $W_1\in\mathbb{R}^{1536\times256}$ and $W_2\in\mathbb{R}^{1536\times1536}$. Each transformer block in the action expert then produces scale and shift vectors:
\begin{align}
    [s_\ell;b_\ell]
        &= W_\ell\,\mathrm{SiLU}(c_\tau)+d_\ell, \\
    z_\ell
        &= (1+s_\ell)\odot\mathrm{LN}(h_\ell)+b_\ell.
\end{align}
GR00T's LayerNorm, unlike $\pi_{0.5}$'s RMS normalization, subtracts the feature mean before normalization and does not produce a timestep-dependent residual gate. Each block-level projection has shape $3072\times1536$, consisting of equal-size $1536\times1536$ blocks for its two outputs, the scale and shift vectors. Together, the Time MLP and adaptive LayerNorm contain $83.02$M and $158.57$M parameters in N1.5 and N1.6, respectively.

\paragraph{Action--timestep encoder.}
GR00T additionally uses an action--timestep encoder to combine timestep information with noisy actions before the transformer:
\begin{align}
    e_a &= A_1a_\tau+q_1, \\
    u_\tau
        &= \mathrm{SiLU}\!\left(
            A_2[e_a;\phi_a(\tau)]+q_2
        \right), \\
    e_{\mathrm{action}}
        &= A_3u_\tau+q_3.
\end{align}
Importantly, these projections are embodiment-specific. As $e_a$ and $\phi_a(\tau)$ each have dimension $1536$, $A_2\in\mathbb{R}^{1536\times3072}$ for each embodiment. We partition it as:
\begin{equation}
    A_2=[A_{2,\mathrm{action}},A_{2,\mathrm{time}}],
\end{equation}
where each block has shape $1536\times1536$. We include $A_{2,\mathrm{time}}$ in the Timestep Modules, but exclude $A_{2,\mathrm{action}}$ and the subsequent joint projection $A_3$. The timestep-input blocks contain $32\times1536\times1536=75.50$M parameters across the stored embodiment-specific encoders. Overall, the Timestep Modules contain $158.52$M parameters in N1.5 and $234.07$M in N1.6, accounting for $14.83\%$ and $16.50\%$ of their complete action experts, respectively. Notably, while the adaptive-normalization vectors depend only on the timestep, the action-token embeddings depend jointly on the timestep and the noisy action.

\subsection{SmolVLA: Action--Timestep MLP}
SmolVLA uses a different architecture, with an action--timestep MLP rather than adaptive normalization. During inference, each noisy action is projected to an embedding, concatenated with a sinusoidal timestep embedding, and passed through a two-layer MLP:
\begin{align}
    e_a &= W_a a_\tau+b_a, \\
    e_{\mathrm{action}}
        &= W_{\mathrm{out}}\,
        \mathrm{SiLU}\!\left(
            W_{\mathrm{in}}[e_a;\phi(\tau)]+b_{\mathrm{in}}
        \right)+b_{\mathrm{out}}.
\end{align}
The resulting embeddings simply serve as input tokens to the action expert, and thus SmolVLA does not generate timestep-dependent scale, shift, or gate vectors at each transformer layer. The action and timestep embeddings for SmolVLA each have dimension $720$. Thus, $W_{\mathrm{in}}\in\mathbb{R}^{720\times1440}$ and $W_{\mathrm{out}}\in\mathbb{R}^{720\times720}$.
We partition the input matrix as:
\begin{equation}
    W_{\mathrm{in}}
    =
    [W_{\mathrm{action}},W_{\mathrm{time}}],
\end{equation}
and separately analyze the two $720\times720$ blocks. As with GR00T, we include only $W_{\mathrm{time}}$ in the Timestep Module. Overall, the timestep-input block contains $0.52$M parameters, accounting for only $0.52\%$ of the action expert.

\clearpage
\section{Experiment Settings}
\label{app:exp_settings}
This section details the experiment settings used. All experiments used 4 NVIDIA H100 80GB HBM3 GPUs paired with an Intel Xeon Platinum 8480+.

\subsection{Model List}
\label{app:model_list}
We use both publicly released and in-house-trained checkpoints from multiple flow-based VLA model families. Table~\ref{tab:checkpoints} lists all public checkpoints and their comparison references, grouped by architecture, with their Hugging Face links. We use the official OpenPI release for the $\pi_{0.5}$ base checkpoint, distributed at \texttt{gs://openpi-assets/checkpoints/pi05\_base}. All other $\pi_{0.5}$ and GR00T checkpoints are released by RLinf~\citep{zang2025rlinf, yu2026rlinf}. We use Flow-SDE-based $\pi_{0.5}$ RL checkpoints for MetaWorld, ManiSkill, and CALVIN. All RL checkpoints are trained with PPO, except for SmolVLA, which uses GRPO~\citep{shao2024deepseekmath}.

\begin{table}[!htb]
    \centering
    \small
    \renewcommand{\arraystretch}{1.15}
    \setlength{\tabcolsep}{5pt}
    \caption{
        \textbf{Public checkpoints and comparison references.}
    }
    \label{tab:checkpoints}
    \begin{tabularx}{\linewidth}{@{}llXX@{}}
        \toprule
        Update & Benchmark & Reference & Post-trained checkpoint \\
        \midrule

        % ---------------- pi_0.5 ----------------
        \multicolumn{4}{@{}l}{\textbf{$\boldsymbol{\pi_{0.5}}$}} \\
        BC & LIBERO
           & OpenPI $\pi_{0.5}$ base
           & \hfmodel{RLinf/RLinf-Pi05-LIBERO-SFT}{LIBERO few-shot SFT} \\
        BC & MetaWorld
           & OpenPI $\pi_{0.5}$ base
           & \hfmodel{RLinf/RLinf-Pi05-MetaWorld-SFT}{MetaWorld SFT} \\
        BC & ManiSkill
           & OpenPI $\pi_{0.5}$ base
           & \hfmodel{RLinf/RLinf-Pi05-ManiSkill-25Main-SFT}{ManiSkill-25Main SFT} \\
        BC & CALVIN
           & OpenPI $\pi_{0.5}$ base
           & \hfmodel{RLinf/RLinf-Pi05-CALVIN-ABC-D-SFT}{CALVIN ABC-D SFT} \\
        BC & LIBERO
           & \hfmodel{RLinf/RLinf-Pi05-LIBERO-SFT}{LIBERO few-shot SFT}
           & \hfmodel{RLinf/RLinf-Pi05-LIBERO-130-fullshot-SFT}{LIBERO full-shot SFT} \\
        \addlinespace[2pt]
        RL & MetaWorld
           & \hfmodel{RLinf/RLinf-Pi05-MetaWorld-SFT}{MetaWorld SFT}
           & \hfmodel{RLinf/RLinf-Pi05-MetaWorld-RL-FlowSDE}{MetaWorld RL-FlowSDE} \\
        RL & ManiSkill
           & \hfmodel{RLinf/RLinf-Pi05-ManiSkill-25Main-SFT}{ManiSkill-25Main SFT}
           & \hfmodel{RLinf/RLinf-Pi05-ManiSkill-25Main-RL-FlowSDE}{ManiSkill-25Main RL-FlowSDE} \\
        RL & CALVIN
           & \hfmodel{RLinf/RLinf-Pi05-CALVIN-ABC-D-SFT}{CALVIN ABC-D SFT}
           & \hfmodel{RLinf/RLinf-Pi05-CALVIN-ABC-D-RL-FlowSDE}{CALVIN ABC-D RL-FlowSDE} \\
        \midrule

        % ---------------- GR00T N1.5 ----------------
        \multicolumn{4}{@{}l}{\textbf{GR00T N1.5}} \\
        RL & LIBERO-Spatial
           & \hfmodel{RLinf/RLinf-Gr00t-SFT-Spatial}{Spatial SFT}
           & \hfmodel{RLinf/RLinf-Gr00t-RL-Spatial-Step400}{Spatial RL, step 400} \\
        RL & LIBERO-Object
           & \hfmodel{RLinf/RLinf-Gr00t-SFT-Object}{Object SFT}
           & \hfmodel{RLinf/RLinf-Gr00t-RL-Object-Step400}{Object RL, step 400} \\
        RL & LIBERO-Goal
           & \hfmodel{RLinf/RLinf-Gr00t-SFT-Goal}{Goal SFT}
           & \hfmodel{RLinf/RLinf-Gr00t-RL-Goal-Step500}{Goal RL, step 500} \\
        \midrule

        % ---------------- GR00T N1.6 ----------------
        \multicolumn{4}{@{}l}{\textbf{GR00T N1.6}} \\
        RL & LIBERO-Spatial
           & \hfmodel{RLinf/RLinf-Gr00t-N1.6-SFT-Spatial}{Spatial SFT}
           & \hfmodel{RLinf/RLinf-Gr00t-N1.6-RL-Spatial-Step500}{Spatial RL, step 500} \\
        \midrule

        % ---------------- SmolVLA ----------------
        \multicolumn{4}{@{}l}{\textbf{SmolVLA} (community releases)} \\
        BC & LIBERO-Spatial
           & \hfmodel{lerobot/smolvla_base}{SmolVLA base}
           & \hfmodel{katsukiono/smolvla-libero-spatial-4arm/tree/main/sft}{Spatial SFT} \\
        BC & LIBERO-Spatial
           & \hfmodel{katsukiono/smolvla-libero-spatial-4arm/tree/main/sft}{Spatial SFT}
           & \hfmodel{katsukiono/smolvla-libero-spatial-4arm/tree/main/rs_sft}{Spatial RS-SFT} \\
        \addlinespace[2pt]
        RL & LIBERO-Spatial
           & \hfmodel{katsukiono/smolvla-libero-spatial-4arm/tree/main/sft}{Spatial SFT}
           & \hfmodel{katsukiono/smolvla-libero-spatial-4arm/tree/main/grpo}{Spatial GRPO} \\
        RL & LIBERO-Object
           & \hfmodel{MorpheusTzz/smolvla-grpo-libero-object/tree/main/sft-100pct-baseline}{Object SFT}
           & \hfmodel{MorpheusTzz/smolvla-grpo-libero-object/tree/main/grpo-seed11-update300}{Object GRPO, update 300} \\

        \bottomrule
    \end{tabularx}
\end{table}

\subsection{In-house Training Setup}
We also train both BC and RL policies in-house. Specifically, we train (1) $\pi_{0.5}$ RL on LIBERO-Spatial, LIBERO-Object, and LIBERO-Goal, as the official RLinf checkpoints are unavailable, (2) GR00T N1.5 and GR00T N1.6 RL on LIBERO for a few steps, as the official checkpoints have a floating-point mismatch (\S~\ref{app:vla_models_parameter_updates}), and (3) $\pi_{0.5}$ BC on LIBERO-Spatial and LIBERO for task-controlled BC training and the discrete-timestep ablation (\S~\ref{sec:analysis_rl_induced_update}). For all $\pi_{0.5}$ training, we initialize from the public \hfmodel{RLinf/RLinf-Pi05-LIBERO-SFT}{LIBERO SFT policy}, and for GR00T training, we initialize from the respective reference checkpoints in Table~\ref{tab:checkpoints}.

We detail our RL training settings in Table~\ref{tab:rl_hyperparams}. Following $\pi_{\texttt{RL}}$~\citep{chen2025pi_rl}, we use PPO as our base RL algorithm and adopt Flow-SDE for exploration. We mostly follow the recommended hyperparameters, except for $\pi_{0.5}$, where we found 240 interaction steps to be sufficient for LIBERO-Object and $n=5$ denoising steps to work better for LIBERO-Spatial. We detail our BC hyperparameter settings in Table~\ref{tab:bc_hyperparams}.

\begin{table}[!hbt]
    \centering
    \caption{\textbf{In-house RL hyperparameters on LIBERO.}}
    \label{tab:rl_hyperparams}
    \small
    \renewcommand{\arraystretch}{1.12}
    \setlength{\tabcolsep}{3pt}
    \begin{tabular}{@{}lccccccc@{}}
    \toprule
    & \multicolumn{3}{c}{\textbf{$\boldsymbol{\pi}_{0.5}$}}
    & \multicolumn{3}{c}{\textbf{GR00T N1.5}}
    & \textbf{GR00T N1.6} \\
    \cmidrule(lr){2-4}\cmidrule(lr){5-7}\cmidrule(lr){8-8}
    \textbf{Hyperparameter} & \textbf{Spatial} & \textbf{Object} & \textbf{Goal} & \textbf{Spatial} & \textbf{Object} & \textbf{Goal} & \textbf{Spatial} \\
    \midrule
    \multicolumn{8}{@{}l}{\textit{Optimization}} \\
    Max steps & 150 & 120 & 150 & 25 & 5 & 5 & 5 \\
    Global batch size & 2048 & 2048 & 2048 & 1024 & 768 & 1024 & 720 \\
    Update epochs & 1 & 1 & 4 & 4 & 4 & 4 & 4 \\
    Actor learning rate & $5\times10^{-6}$ & $5\times10^{-6}$ & $5\times10^{-6}$ & $5\times10^{-6}$ & $5\times10^{-6}$ & $5\times10^{-6}$ & $5\times10^{-6}$ \\
    Critic learning rate & $1\times10^{-4}$ & $1\times10^{-4}$ & $1\times10^{-4}$ & $1\times10^{-4}$ & $1\times10^{-4}$ & $1\times10^{-4}$ & $1\times10^{-4}$ \\
    \midrule
    \multicolumn{8}{@{}l}{\textit{PPO}} \\
    Discount factor $\gamma$ & 0.99 & 0.99 & 0.99 & 0.99 & 0.99 & 0.99 & 0.99 \\
    GAE $\lambda$ & 0.95 & 0.95 & 0.95 & 0.95 & 0.95 & 0.95 & 0.95 \\
    PPO clip ratio $\epsilon$ & 0.2 & 0.2 & 0.2 & 0.2 & 0.2 & 0.2 & 0.2 \\
    KL coefficient & 0 & 0 & 0 & 0 & 0 & 0 & 0.01 \\
    Environment reward scale & 1 & 1 & 1 & 1 & 1 & 1 & 5 \\
    \midrule
    \multicolumn{8}{@{}l}{\textit{Rollout}} \\
    Interaction steps & 240 & 240 & 320 & 240 & 240 & 240 & 240 \\
    Parallel environments & 64 & 64 & 64 & 64 & 48 & 64 & 48 \\
    Rollout epochs & 8 & 8 & 8 & 8 & 8 & 8 & 8 \\
    Action horizon $H$ & 10 & 10 & 10 & 16 & 16 & 16 & 50 \\
    Replan horizon $H'$ & 5 & 5 & 5 & 5 & 5 & 5 & 16 \\
    \midrule
    \multicolumn{8}{@{}l}{\textit{Flow Policy}} \\
    Denoising steps & 5 & 5 & 5 & 4 & 4 & 4 & 4 \\
    Flow-SDE noise $\sigma$ & 0.5 & 0.3 & 0.3 & 0.5 & 0.5 & 0.5 & 0.5 \\
    \bottomrule
    \end{tabular}
\end{table}

\begin{table}[!htb]
    \centering
    \caption{\textbf{In-house BC hyperparameters for \boldmath$\pi_{0.5}$.}}
    \label{tab:bc_hyperparams}
    \small
    \renewcommand{\arraystretch}{1.12}
    \setlength{\tabcolsep}{5pt}
    \begin{tabular}{@{}lc@{}}
    \toprule
    \textbf{Hyperparameter} & \textbf{LIBERO} \\
    \midrule
    Optimizer steps & 1800 \\
    Global batch size & 2048 \\
    Learning rate & $5\times10^{-6}$ \\
    Weight decay & 0.01 \\
    Gradient clipping norm & 1.0 \\
    Action horizon $H$ & 10 \\
    Timestep sampling & Continuous beta \\
    \bottomrule
    \end{tabular}
\end{table}

\clearpage
\section{Extended Results}
This section details the additional experiment results for each sections.

\subsection{Threshold Ablations on Parameter Shift Analyses}
\label{app:threshold_ablations}

We examine whether our findings depend on the thresholds used to measure update density and effective rank. For \textit{density threshold ablations}, we vary the absolute-change threshold from $10^{-7}$ to $10^{-3}$ using LIBERO-Spatial BC and RL checkpoints initialized from the same SFT policy and trained for 600 optimizer updates. For \textit{rank threshold ablations}, we vary the explained-energy threshold over $90\%$, $95\%$, and $99\%$, using the released BC and RL checkpoint comparisons on MetaWorld, ManiSkill, and CALVIN from the main analysis.

\begin{figure}[!htb]
    \centering
    \includegraphics[width=\linewidth]
    {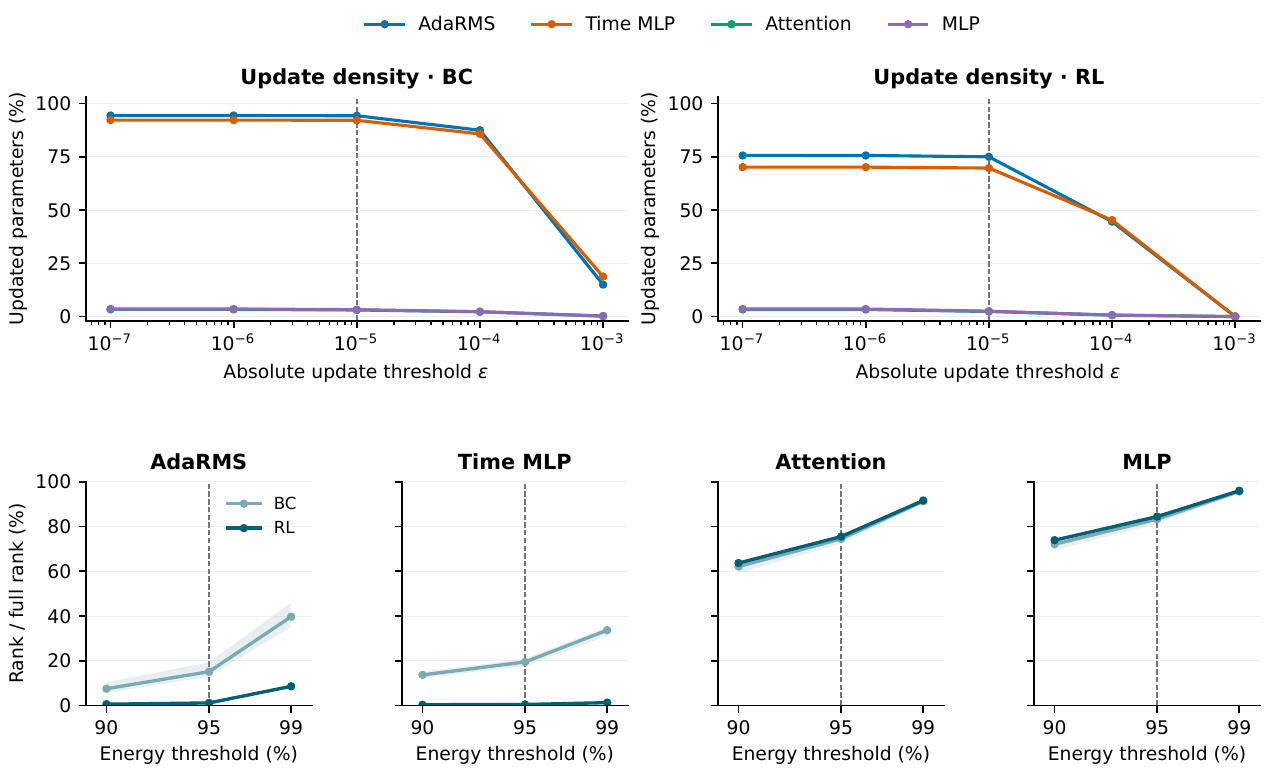}
    \caption{
        \textbf{Sensitivity to measurement thresholds.}
        (Top) Update density across absolute-change thresholds.
        (Bottom) Effective update rank across explained-energy thresholds.
    }
    \label{fig:update-threshold-ablation}
\end{figure}

We find that, similar to Figure~\ref{fig:pi05-update-density-rank}, the Timestep Modules are updated much more densely than attention and MLP modules under both BC and RL across thresholds from $10^{-7}$ to $10^{-4}$ (Figure~\ref{fig:update-threshold-ablation}, top). At $10^{-3}$, however, nearly all RL updates fall below the threshold.

We next examine effective update rank and find that the BC--RL difference in the Timestep Modules persists across all tested energy thresholds (Figure~\ref{fig:update-threshold-ablation}, bottom). Even at $99\%$, AdaRMS updates require only $8.2$--$9.3\%$ of the available rank under RL, compared with $35.3$--$46.0\%$ under BC. In contrast, attention and MLP updates have similarly high ranks under both objectives.

\subsection{Parameter Updates on Flow-Based VLA Models}
\label{app:vla_models_parameter_updates}

We extend our analysis to GR00T~N1.5, GR00T~N1.6, and SmolVLA~\citep{bjorck2025gr00t,smolvla} on LIBERO, under the same settings as the analyses in \S~\ref{sec:analysis_rl_induced_update}. For \textit{GR00T}, we analyze N1.5 RL checkpoints on Spatial, Object, and Goal, and an N1.6 checkpoint on Spatial. For \textit{SmolVLA}, we analyze Spatial BC and RL checkpoints and an Object RL checkpoint. We separate the timestep-input block of the action--timestep MLP from its output matrix.

\begin{figure}[!htb]
    \centering
    \includegraphics[width=\linewidth]
    {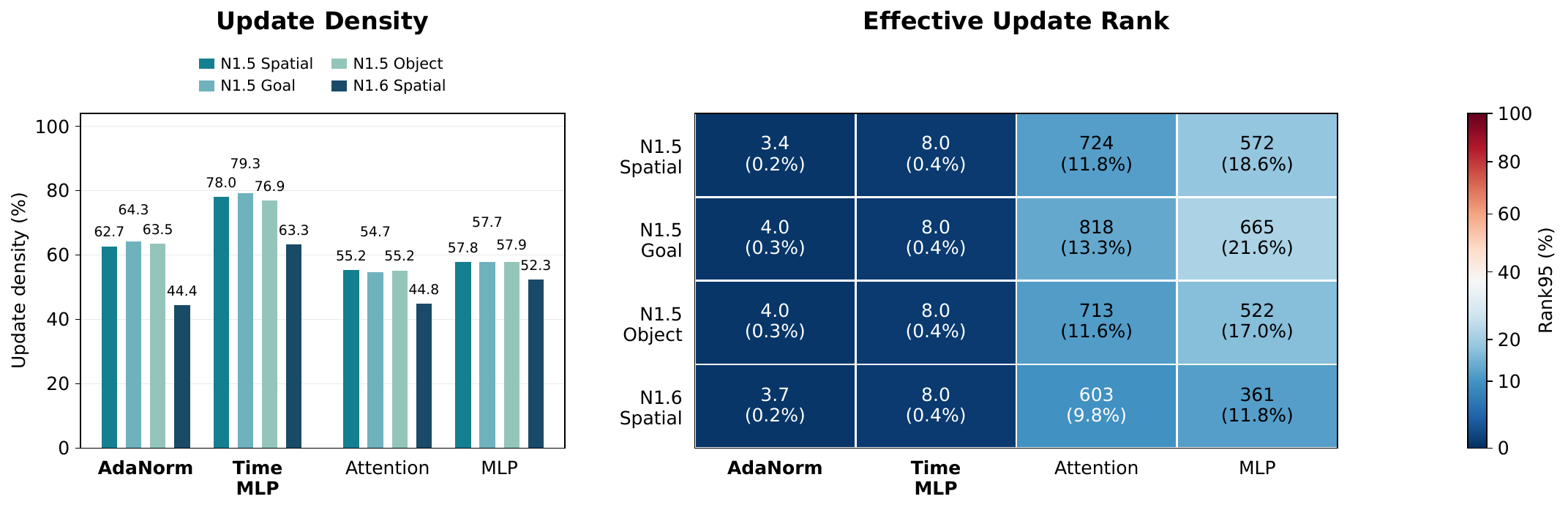}
    \caption{
        \textbf{GR00T RL updates on LIBERO.}
        (Left) Update density. (Right) Effective update rank.
        Parentheses indicate percentages of maximum rank.
    }
    \label{fig:gr00t-update-density-rank}
\end{figure}

We find that GR00T RL updates have low effective ranks in AdaNorm and Time MLP, requiring only $3$--$4$ directions per matrix, while attention and MLP updates require $9.8$--$21.6\%$ of the available rank (Figure~\ref{fig:gr00t-update-density-rank}). All four components show substantial update densities.

\begin{figure}[!htb]
    \centering
    \includegraphics[width=\linewidth]
    {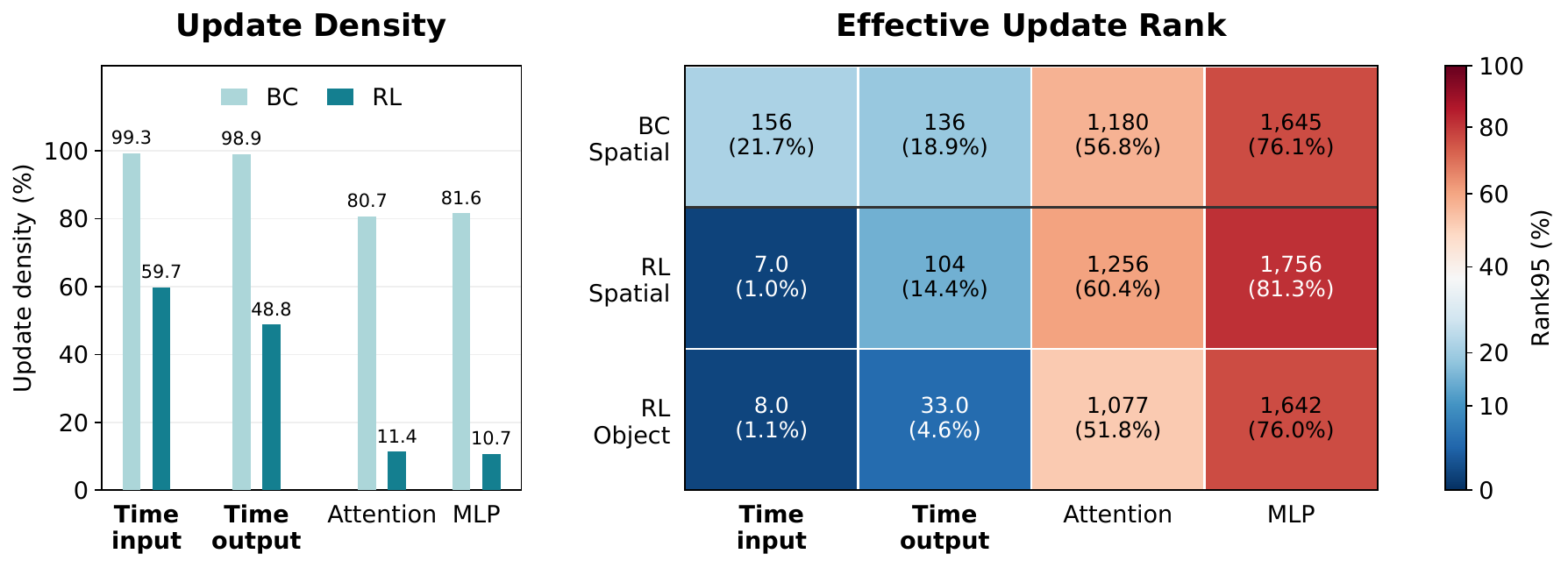}
    \caption{
        \textbf{SmolVLA updates on LIBERO.}
        (Left) Spatial BC and RL update density.
        (Right) Effective update rank, including Object RL.
        Parentheses indicate percentages of maximum rank.
    }
    \label{fig:smolvla-update-density-rank}
\end{figure}

We find a similar pattern in SmolVLA's timestep-input block: RL updates require only $7$--$8$ directions, compared with $156$ under BC (Figure~\ref{fig:smolvla-update-density-rank}). The output matrix requires more directions than the timestep-input block, while attention and MLP updates remain high-rank.

\clearpage

\subsection{Parameter Updates in Image and Video DiT Models}
\label{app:image_video_parameter_analyses}

We extend our parameter analysis to publicly released RL-trained image and video models to examine whether the low-rank updates observed in VLA Timestep Modules also appear in other diffusion- or flow-based models. We analyze FLUX with DanceGRPO~\citep{dancegrpo} and Pref-GRPO~\citep{prefgrpo}, SeFi-Image with DiffusionNFT~\citep{diffusionnft}, and Wan2.2 with VideoRLVR~\citep{videorlvr} (Table~\ref{tab:image-video-checkpoints}). We group the analyzed matrices into similar timestep-related modules, such as AdaNorm and Time MLP, along with attention and MLP modules, with AdaNorm covering architecture-specific modulation projections.

\begin{figure}[!htb]
    \centering
    \includegraphics[width=\linewidth]
    {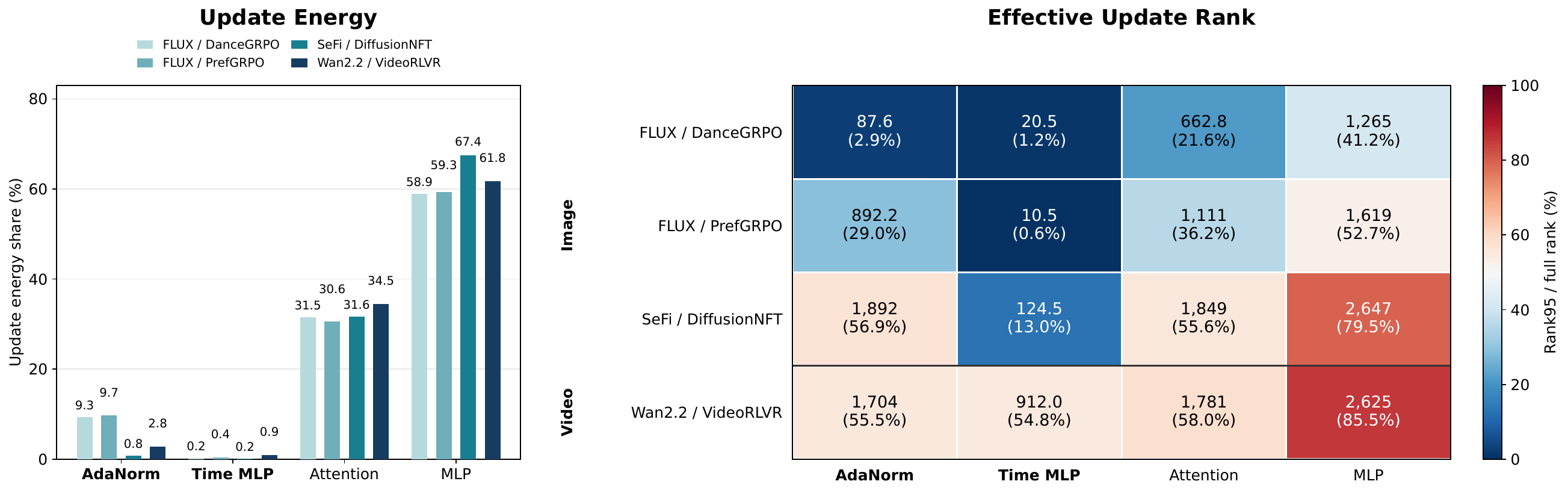}
    \caption{
        \textbf{RL updates in image and video models.}
        (Left) Update-energy share.
        (Right) Mean per-matrix Rank95 and normalized rank.
    }
    \label{fig:image-video-update-rank}
\end{figure}

We find that attention and MLP account for approximately $90$--$99\%$ of the measured update energy (Figure~\ref{fig:image-video-update-rank}, left). Effective ranks vary across models, with AdaNorm updates being relatively high-rank in SeFi-Image and VideoRLVR but lower-rank in DanceGRPO and Pref-GRPO. Time MLP updates also show substantial variation, including low-rank updates in image models (Figure~\ref{fig:image-video-update-rank}, right).

\begin{table}[!htb]
    \centering
    \scriptsize
    \setlength{\tabcolsep}{3pt}
    \caption{\textbf{Reference and RL checkpoint pairs for image and video parameter analyses.}}
    \begin{tabular}{@{}lp{0.35\linewidth}p{0.39\linewidth}@{}}
    \toprule
    Model & Reference & RL checkpoint \\
    \midrule
    DanceGRPO & \href{https://huggingface.co/black-forest-labs/FLUX.1-dev}{FLUX.1-dev} & \href{https://huggingface.co/xzyhku/flux_hpsv2.1_dancegrpo}{xzyhku/flux\_hpsv2.1\_dancegrpo}, checkpoint-300-0 \\
    Pref-GRPO & \href{https://huggingface.co/black-forest-labs/FLUX.1-dev}{FLUX.1-dev} & \href{https://huggingface.co/CodeGoat24/FLUX.1-dev-PrefGRPO}{CodeGoat24/FLUX.1-dev-PrefGRPO} \\
    SeFi-Image & \href{https://huggingface.co/SeFi-Image/SeFi-Image-5B-Base-diffusers}{SeFi-Image-5B-Base-diffusers} & \href{https://huggingface.co/SeFi-Image/SeFi-Image-5B-RL-diffusers}{SeFi-Image-5B-RL-diffusers} \\
    VideoRLVR & \href{https://huggingface.co/DarthZhu/VideoRLVR-Wan2.2-Base}{DarthZhu/VideoRLVR-Wan2.2-Base} & \href{https://huggingface.co/DarthZhu/VideoRLVR-Wan2.2}{DarthZhu/VideoRLVR-Wan2.2} \\
    \bottomrule
    \end{tabular}
    \label{tab:image-video-checkpoints}
\end{table}

\clearpage
\subsection{Low-Rank Parameter Replacement}
\label{app:rank_ablation_analyses}
We extend \S~\ref{sec:Timestep_module_analysis__replacement} to test whether low-rank updates carry the RL gains in the Timestep Modules. For each weight matrix in the Timestep Modules, we compute the SVD of its RL update and reconstruct the update using either only the top four singular directions (\textit{Top Singular Directions only}) or the remaining directions (\textit{w/o Top Singular Directions}), with bias parameters unchanged. As shown in Table~\ref{tab:pi05_singular_directions}, the top four directions alone retain nearly all of the performance of \textit{Timestep Modules only} for $\pi_{0.5}$, whereas removing them drops performance below the base policy on five of six benchmarks, showing that the RL gains are largely captured by the low-rank directions. For GR00T (Table~\ref{tab:groot_singular_directions}), although the checkpoints contain high-rank noise from low-precision training, the top directions still carry most of the gain.

\begin{table}[!htbp]
    \centering
    \renewcommand{\arraystretch}{0.95}
    \setlength{\tabcolsep}{5pt}
    \caption{\textbf{Top singular direction parameter replacement for \boldmath$\pi_{0.5}$.} Best results are \textbf{bolded}.}
    \label{tab:pi05_singular_directions}
    \resizebox{\linewidth}{!}{%
        \begin{tabular}{lcccccc}
        \toprule
        & \multicolumn{5}{c}{$\pi_{0.5}$} \\
        \cmidrule(lr){2-7}
        Condition
        & Spatial & Object & Goal & ManiSkill & MetaWorld& CALVIN \\
        \midrule
        Base
        & 85.0 & 95.4 & 82.8 & 42.2 & 42.8 &  61.8 \\
        \hspace{0.5em}\textit{Timestep Modules only}
        & \textbf{94.8} & \textbf{99.0} & \textbf{91.4} & 87.5 & \textbf{69.6} &\textbf{87.0}\\
        \hspace{0.5em} \textit{w/o Top Singular Directions}
        & 79.6 & 51.8 & 78.4 & 53.8 & 22.6 &48.7\\
        \hspace{0.5em} \textit{w/ Top Singular Directions Only}
        & 90.4 & 96.6 & 89.4 & \textbf{88.4} & 66.6&86.7 \\
        
        \bottomrule
    \end{tabular}
    }
\end{table}

\begin{table}[!htbp]
    \centering
    \renewcommand{\arraystretch}{0.95}
    \setlength{\tabcolsep}{6pt}
\caption{\textbf{Top singular direction parameter replacement for GR00T.} Best results are \textbf{bolded}.}
    \label{tab:groot_singular_directions}
    \begin{tabular}{lcccc}
        \toprule
        &
        \multicolumn{3}{c}{GR00T N1.5} &
        \multicolumn{1}{c}{GR00T N1.6} \\
        \cmidrule(lr){2-4}
        \cmidrule(lr){5-5}
        Condition
        & Spatial & Object & Goal & Spatial \\
        \midrule
        Base
        & 49.4 & 62.4 & 51.4 & 75.6 \\
        \hspace{0.5em}\textit{Timestep Modules only}
        & \textbf{52.6} & \textbf{81.6} & 49.8 & \textbf{84.8} \\
        \hspace{0.5em} \textit{w/o Top Singular Directions}
        & 51.0 & 65.2 & 46.2 & 81.2 \\
        \hspace{0.5em} \textit{w/ Top Singular Directions Only}
        & 51.8 & 81.2 & \textbf{51.8} & 84.0 \\
        \bottomrule
    \end{tabular}
\end{table}

\subsection{Targeted LoRA}
\label{app:lora_experiment}
Building on Takeaway~\ref{tk:module_replacement}, we investigate applying LoRA~\citep{hu2022lora} only to the Timestep Modules, especially since RLinf's default LoRA target omits them (Figure~\ref{fig:lora-timestep-targets}). We compare LoRA on the Timestep Modules against the base settings in Table~\ref{tab:lora-spatial-learning}, where targeting the Timestep Modules generally improves performance earlier in training and achieves higher peak and final success.

\begin{figure}[!hbt]
\centering
\begin{minipage}{\linewidth}
\begin{lstlisting}[
    language=Python,
    basicstyle=\small\ttfamily,
    keywordstyle=\bfseries,
    commentstyle=\itshape,
    stringstyle=\ttfamily,
    showstringspaces=false,
    columns=fullflexible,
    keepspaces=true,
    breaklines=true,
    frame=tb,
    rulecolor=\color{black},
    aboveskip=0pt,
    belowskip=0pt,
    framesep=3pt
]
target_modules = [
    "proj", "qkv", "fc1", "fc2", "q", "kv",
    "fc3", "out_proj",
    "q_proj", "k_proj", "v_proj", "o_proj",
    "gate_proj", "up_proj", "down_proj", "lm_head",
]
# Excludes Time MLP and AdaRMS projections.
\end{lstlisting}
\end{minipage}
\caption{\textbf{RLinf's default LoRA targets omit Timestep Modules.}
The $\pi_{0.5}$ Time MLP and AdaRMS projections are absent from
\href{https://github.com/RLinf/RLinf/blob/184ba07f0a5b3a09a4696a4e3c8aa4b3c720fa86/rlinf/models/__init__.py}{RLInf Codebase}.}
\label{fig:lora-timestep-targets}
\end{figure}

\begin{table}[!htb]
    \centering
    \caption{\textbf{Targeted LoRA on LIBERO-Spatial with \boldmath$\pi_{0.5}$.}
    Success rates (\%) over training steps with LoRA, rank=32.
    The higher success rate at each step is \textbf{bolded}.}
    \label{tab:lora-spatial-learning}
    \small
    \renewcommand{\arraystretch}{1.12}
    \setlength{\tabcolsep}{2.5pt}
    \begin{tabular}{@{}l*{11}{c}@{}}
    \toprule
    \textbf{LoRA target}
    & \textbf{0} & \textbf{10} & \textbf{20}
    & \textbf{30} & \textbf{40} & \textbf{50}
    & \textbf{60} & \textbf{70} & \textbf{80}
    & \textbf{90} & \textbf{100} \\
    \midrule
    Time MLP + AdaRMS
    & 85.0 & \textbf{85.6} & \textbf{90.0}
    & \textbf{91.4} & \textbf{91.8} & \textbf{92.0}
    & 90.8 & \textbf{92.0} & \textbf{94.2}
    & \textbf{93.0} & \textbf{93.0} \\
    Attn + FFN
    & 85.0 & 81.0 & 83.4
    & 87.0 & 89.4 & 89.0
    & \textbf{91.6} & 91.4 & 92.0
    & 92.0 & 91.4 \\
    \bottomrule
    \end{tabular}
\end{table}

\clearpage
\subsection{Time Conditioning Directions in AdaRMS Updates}
\label{app:time_conditioning}
We extend the timestep analysis to all 18 action-expert layers, showing the attention and MLP AdaRMS sites separately. For each joint scale--shift--gate update, we plot the top three input singular directions using $|\cos(v_i,c_\tau)|$, normalized by each curve's maximum.

\begin{figure*}[!htb]
  \centering
  \includegraphics[page=1,width=\textwidth]{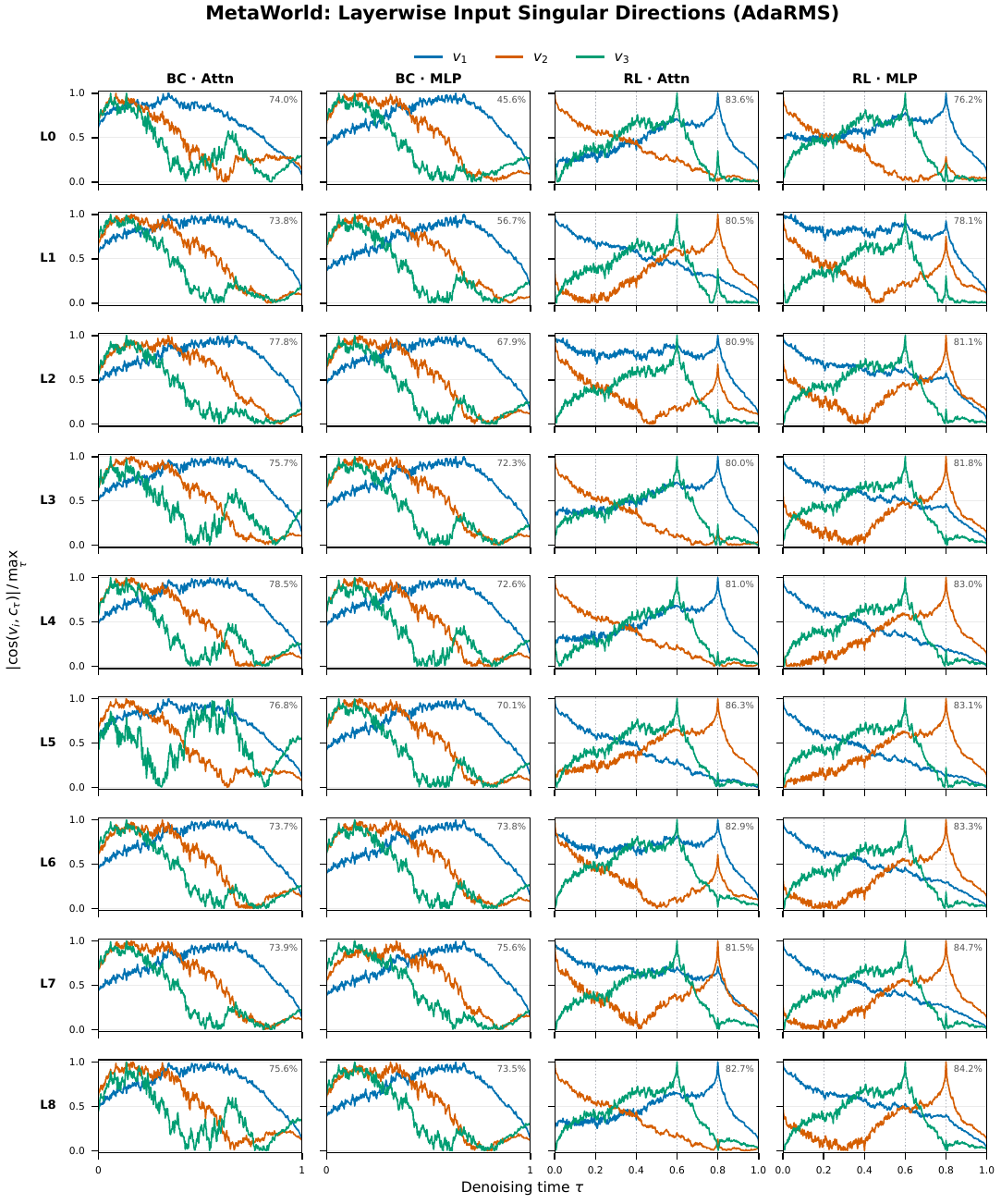}
  \caption{\textbf{Layerwise BC and RL AdaRMS input directions on MetaWorld} (Layers 0--8).}
  \label{fig:metaworld-layerwise-1}
\end{figure*}

\begin{figure*}[!htb]
  \centering
  \includegraphics[page=2,width=\textwidth]{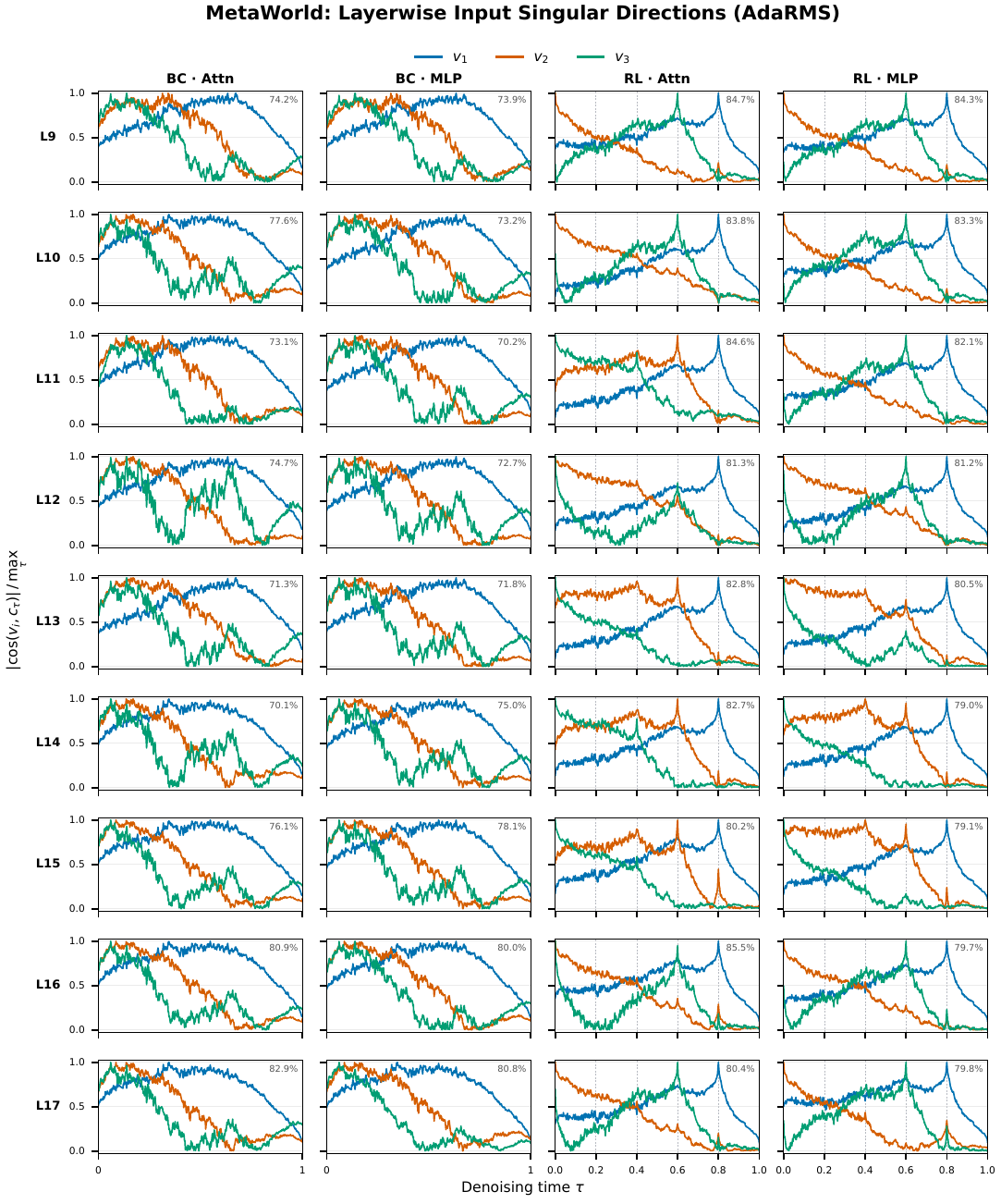}
  \caption{\textbf{Layerwise BC and RL AdaRMS input directions on MetaWorld} (Layers 9--17).}
  \label{fig:metaworld-layerwise-2}
\end{figure*}

\begin{figure*}[!htb]
  \centering
  \includegraphics[page=1,width=\textwidth]{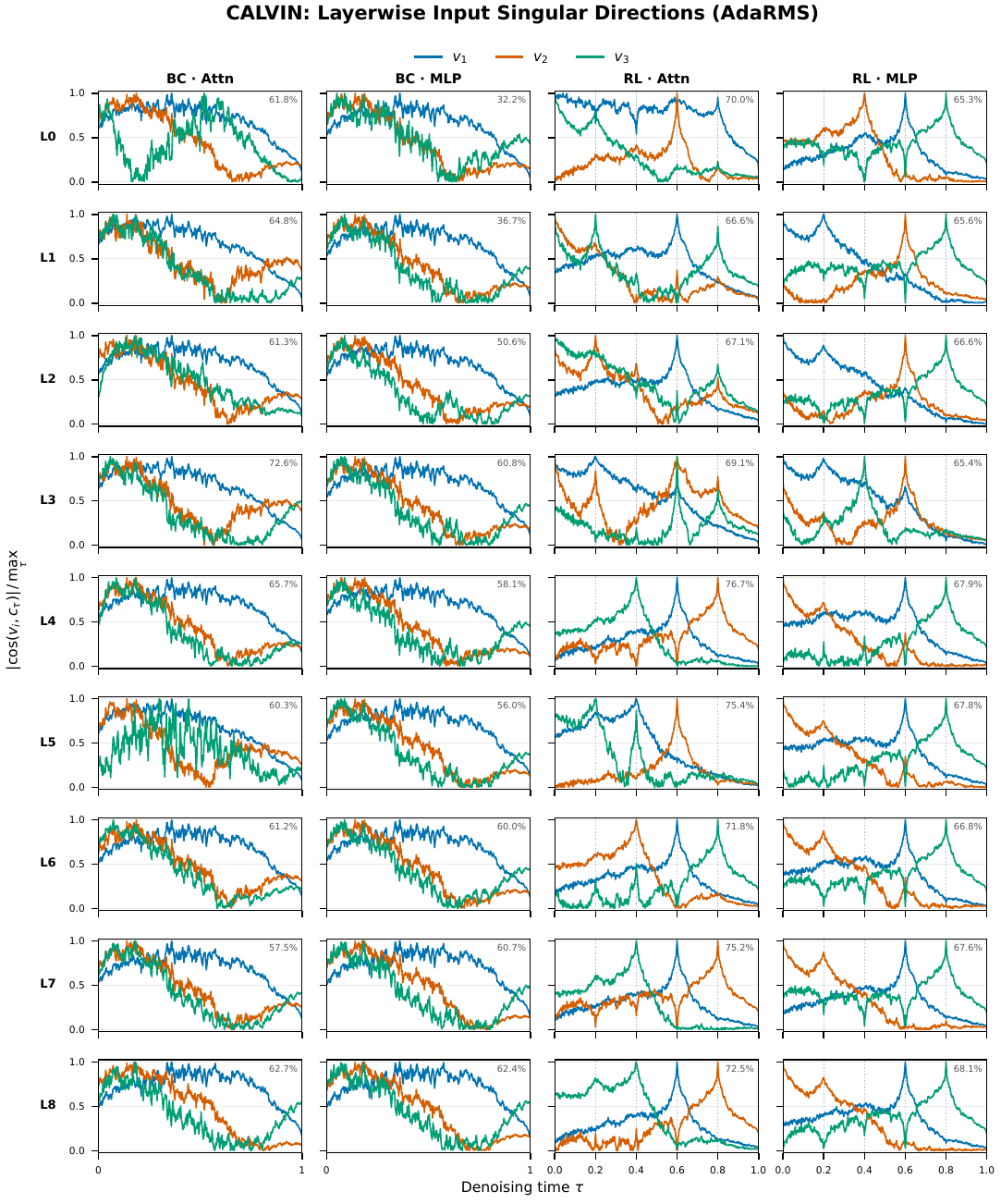}
  \caption{\textbf{Layerwise BC and RL AdaRMS input directions on CALVIN} (Layers 0--8).}
  \label{fig:calvin-layerwise-1}
\end{figure*}

\begin{figure*}[!htb]
  \centering
  \includegraphics[page=2,width=\textwidth]{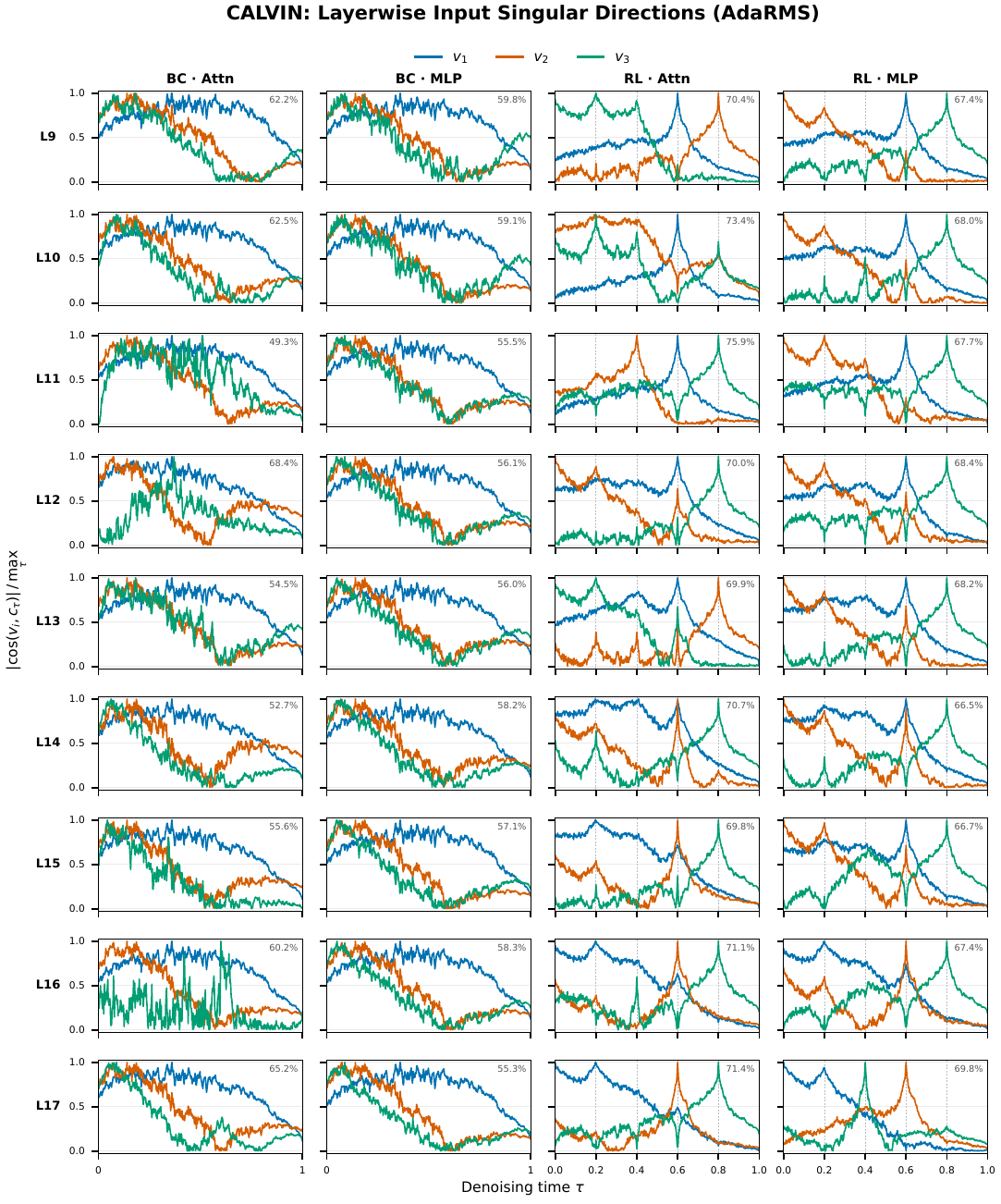}
  \caption{\textbf{Layerwise BC and RL AdaRMS input directions on CALVIN} (Layers 9--17).}
  \label{fig:calvin-layerwise-2}
\end{figure*}

\begin{figure*}[!htb]
  \centering
  \includegraphics[page=1,width=\textwidth]{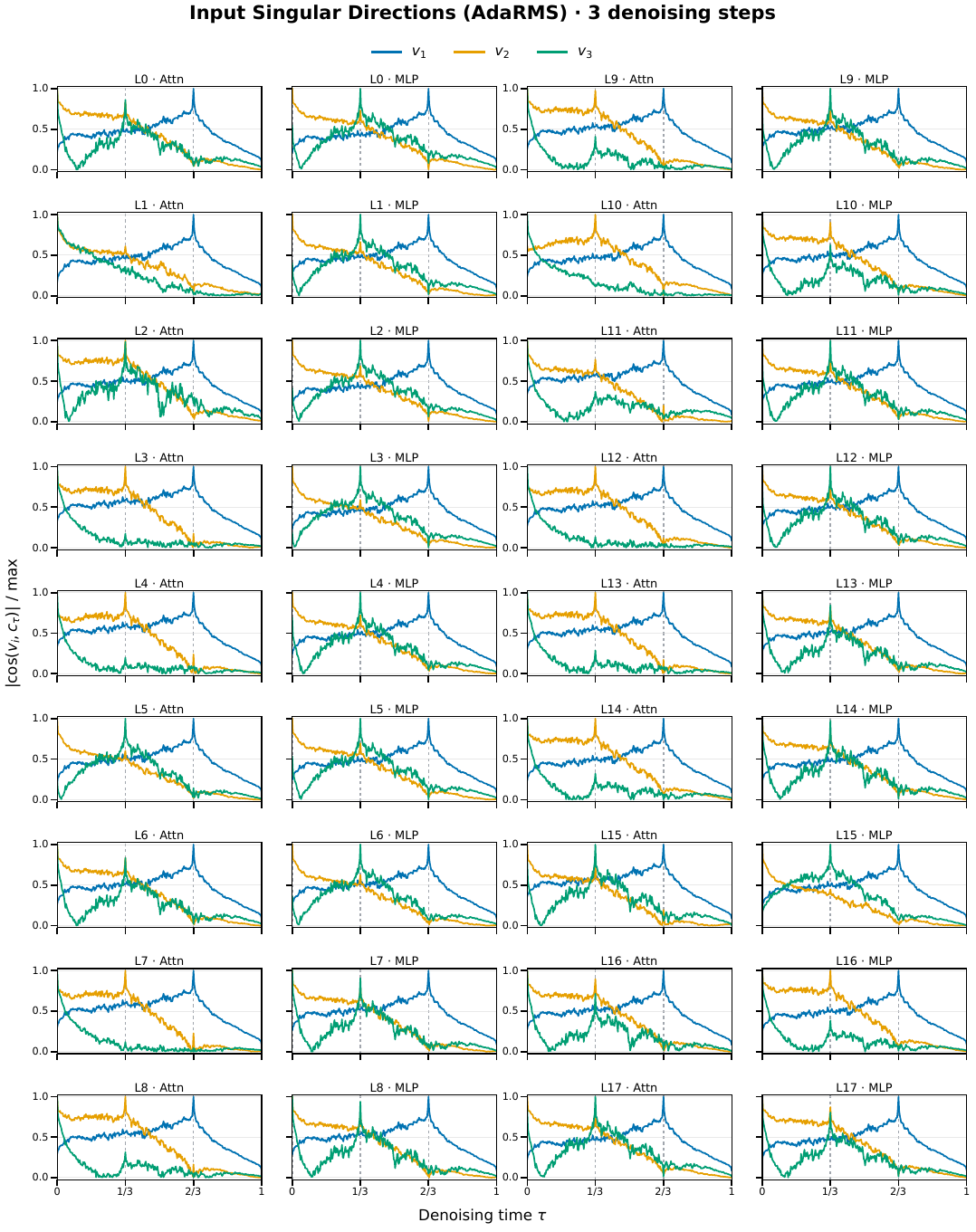}
  \caption{\textbf{Layerwise AdaRMS input directions for LIBERO-Spatial RL with three denoising steps.}}
  \label{fig:spatial-layerwise-3step}
\end{figure*}

\begin{figure*}[!htb]
  \centering
  \includegraphics[page=2,width=\textwidth]{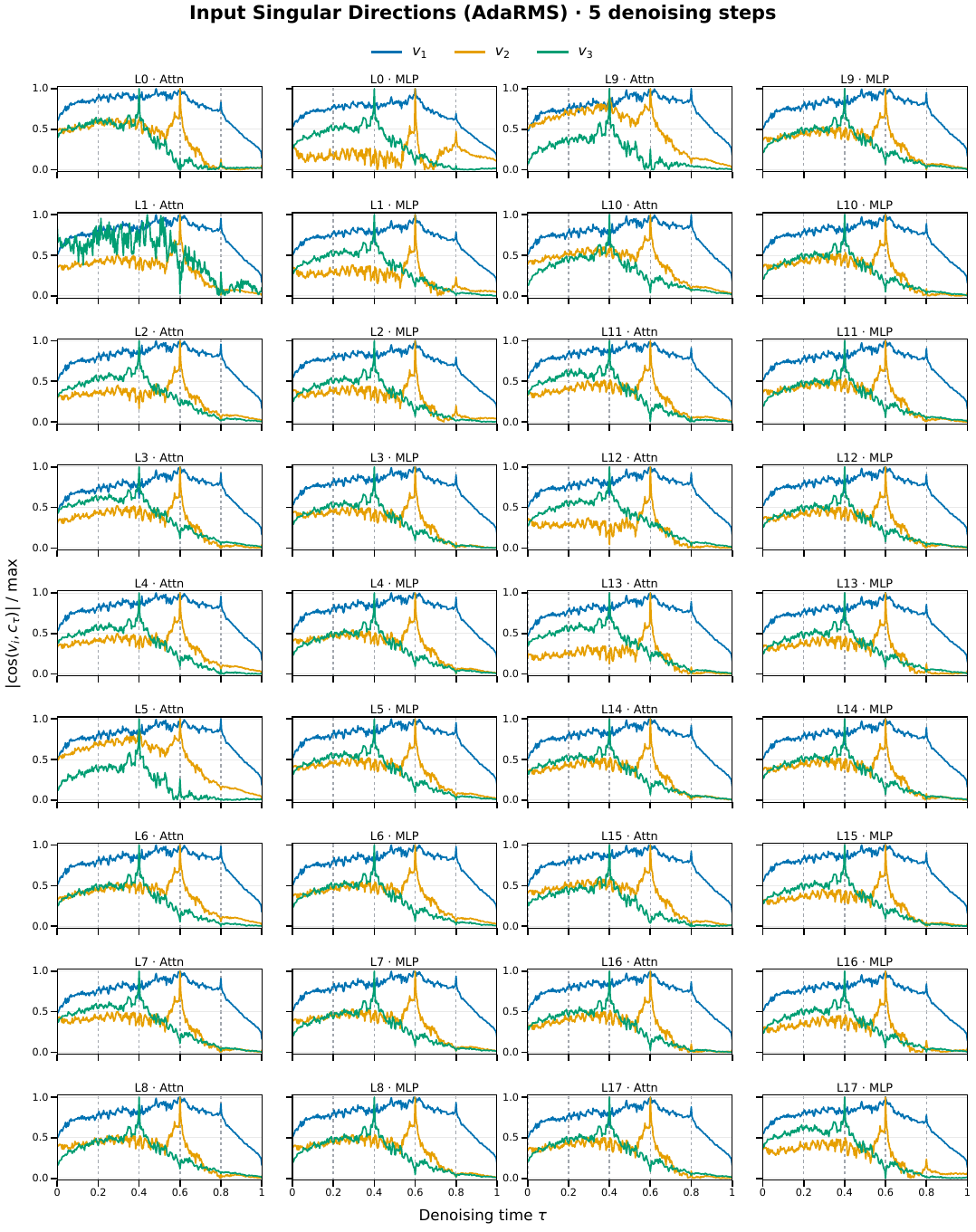}
  \caption{\textbf{Layerwise AdaRMS input directions for LIBERO-Spatial with RL five denoising steps.}}
  \label{fig:spatial-layerwise-5step}
\end{figure*}

\begin{figure}[p]
    \centering
    \includegraphics[page=1,width=\linewidth]{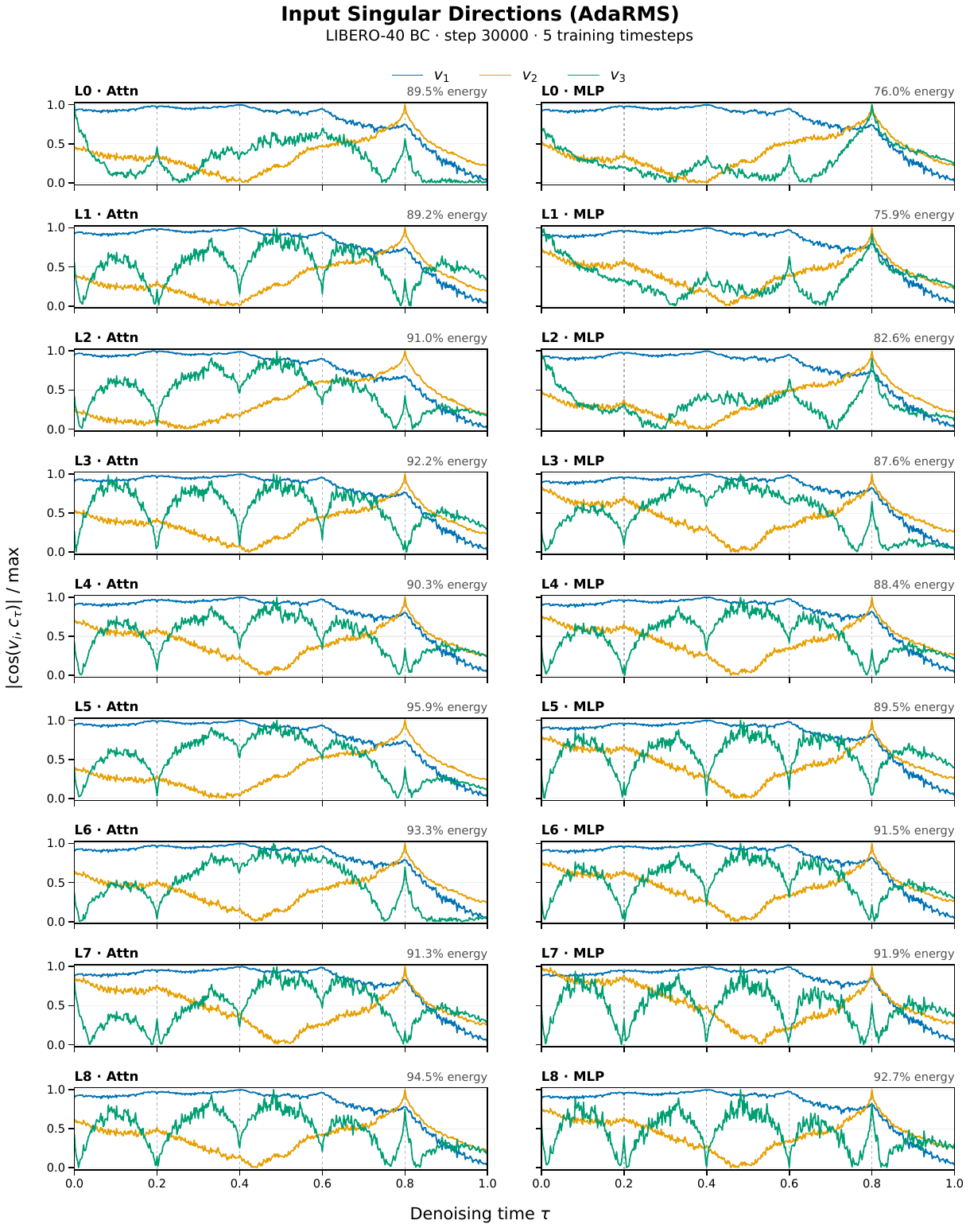}
    \caption{\textbf{Layerwise AdaRMS input directions for BC trained with five discrete timesteps} (Layers 0--8).}
    \label{fig:discrete-bc-layerwise-1}
\end{figure}

\begin{figure}[p]
    \centering
    \includegraphics[page=2,width=\linewidth]{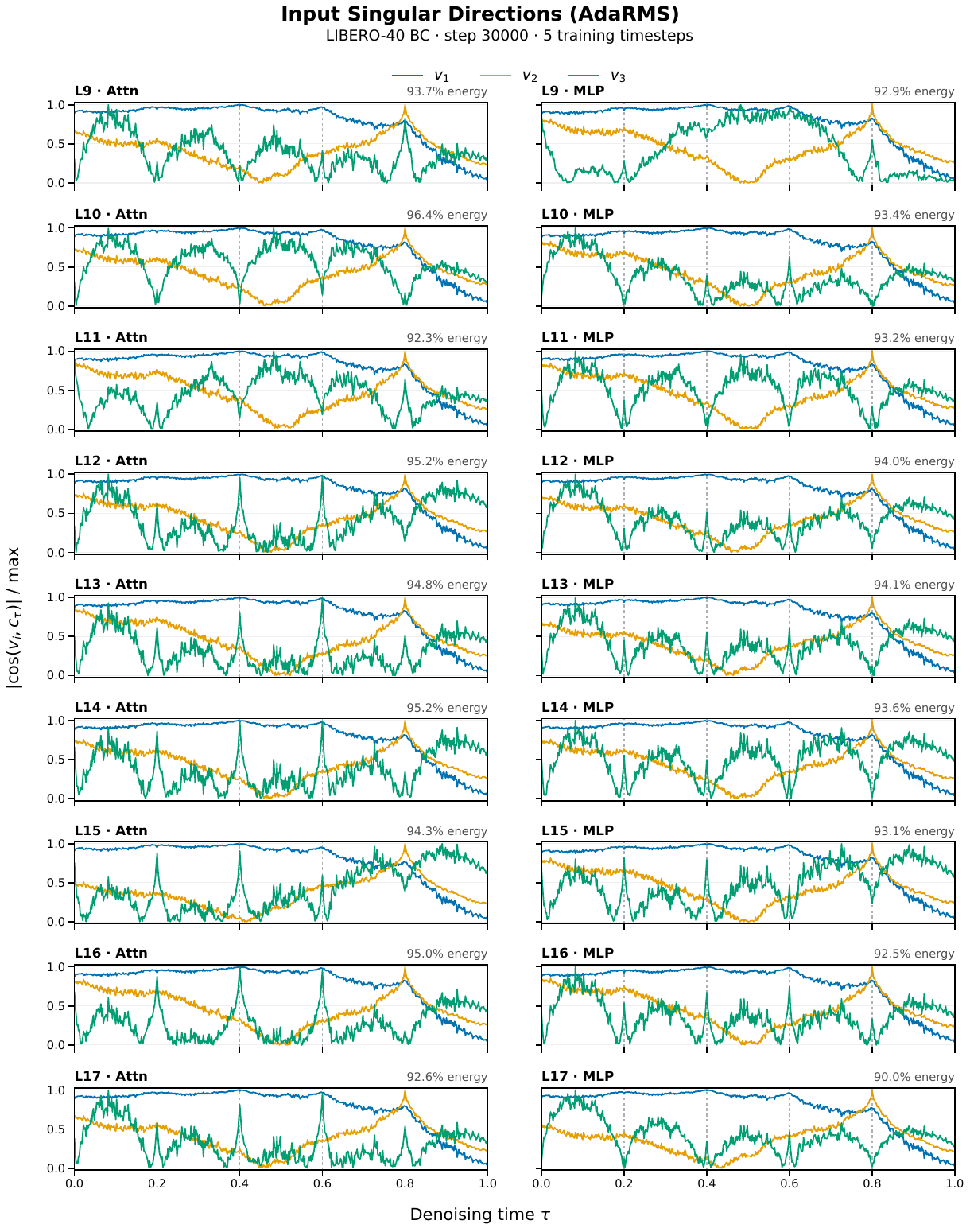}
    \caption{\textbf{Layerwise AdaRMS input directions for BC trained with five discrete timesteps} (Layers 9--17).}
    \label{fig:discrete-bc-layerwise-2}
\end{figure}

Figures~\ref{fig:metaworld-layerwise-1}--\ref{fig:calvin-layerwise-2} compare BC and RL on MetaWorld and CALVIN. Figures~\ref{fig:spatial-layerwise-3step} and~\ref{fig:spatial-layerwise-5step} compare LIBERO-Spatial RL policies trained with three and five denoising steps. Figures~\ref{fig:discrete-bc-layerwise-1} and~\ref{fig:discrete-bc-layerwise-2} show LIBERO-40 BC trained for 30,000 optimizer updates with $\tau \in \{0,0.2,0.4,0.6,0.8\}$.
% MetaWorld

% CALVIN

% LIBERO Spatial: 3 vs. 5 denoising steps

\clearpage

\subsection{Comparing Scale, Shift, and Gate}
\label{app:scale_shift_gate}
\begin{figure*}[!htbp]
  \centering
  \includegraphics[page=1,width=0.9\textwidth]{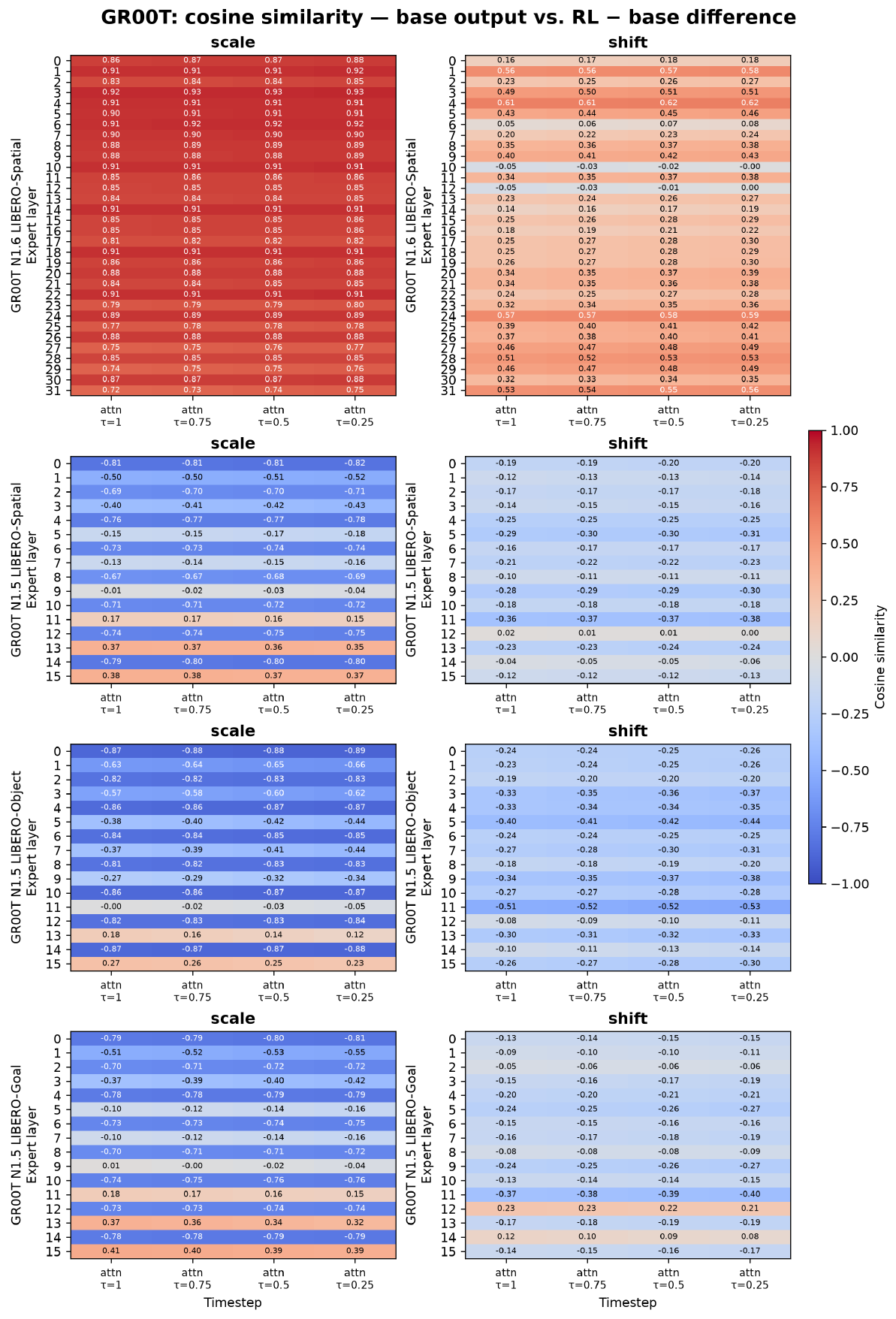}
  \caption{\textbf{Position-wise cosine similarity between each base output and its RL-induced change in GR00T checkpoints.}}
  \label{fig:scale-shift-gate-groot}
\end{figure*}

\begin{figure*}[!htbp]
  \centering
  \includegraphics[page=1,width=0.85\textwidth]{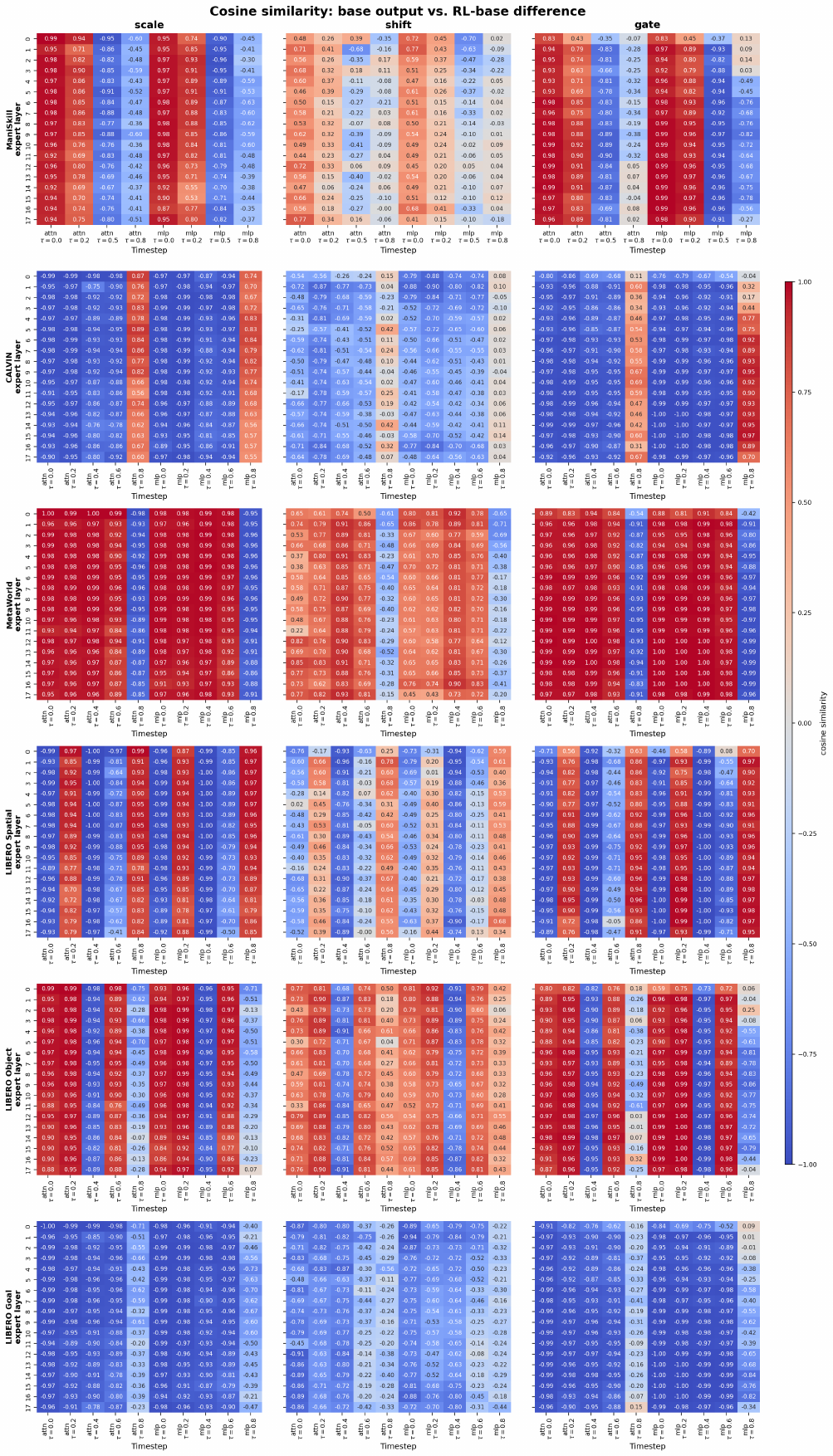}
  \caption{\textbf{Position-wise cosine similarity between each base output and its RL-induced change in \boldmath$\pi_{0.5}$ checkpoints.}}
  \label{fig:scale-shift-gate-pi05}
\end{figure*}

Figures~\ref{fig:scale-shift-gate-groot} and~\ref{fig:scale-shift-gate-pi05} compare RL-induced directional changes in scale, shift, and gate across $\pi_{0.5}$ and GR00T. The shift vector changes more distinctly, motivating our focus on it.

\clearpage
\subsection{Probing Results}
\label{app:probing_results}

\begin{figure*}[!htb]
  \centering
  \includegraphics[width=\linewidth]
  {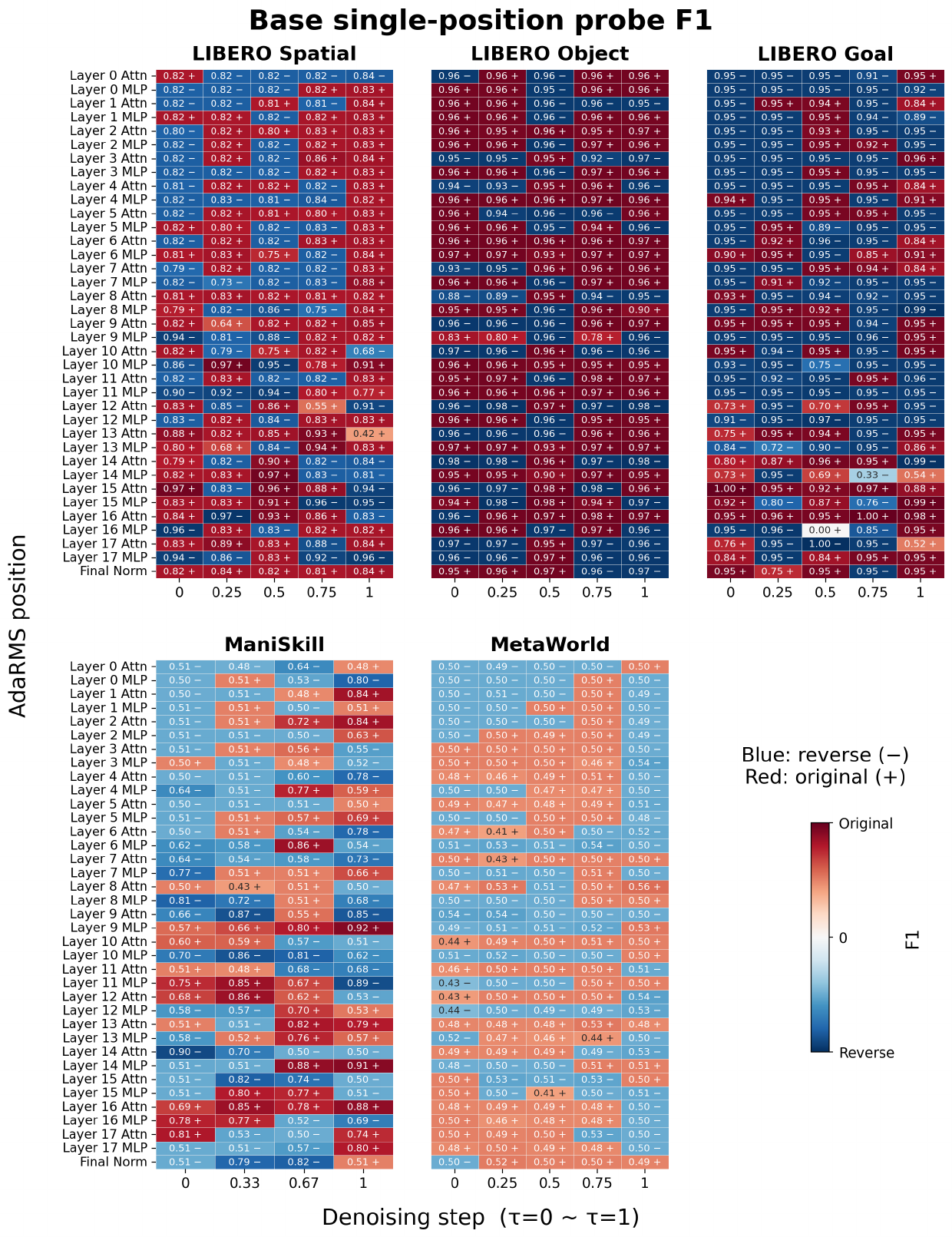}
  \caption{\textbf{Single-position probing results in \boldmath$\pi_{0.5}$ base policies.}
Each sublayer--timestep direction is used individually as a probe. Red indicates that larger projections are associated with success (Original), while blue indicates the opposite (Reverse). Values denote F1 scores on the test split.}
  \label{fig:base_reverse_original}
\end{figure*}

\begin{figure*}[!htb]
  \centering
  \includegraphics[width=\linewidth]
  {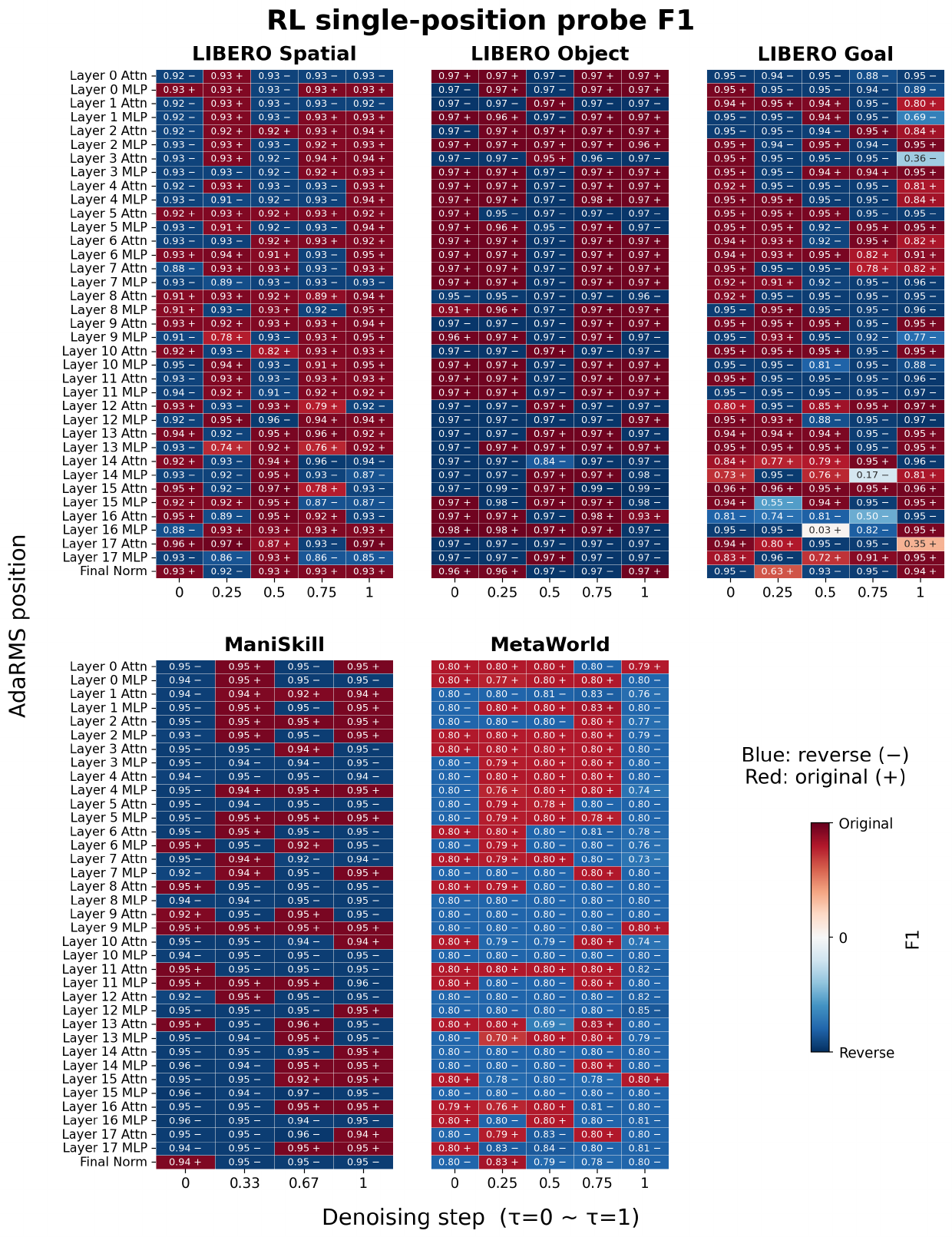}
  \caption{\textbf{Single-position probing results in \boldmath$\pi_{0.5}$ RL policies.}
Each sublayer--timestep direction is used individually as a probe. Red indicates that larger projections are associated with success (Original), while blue indicates the opposite (Reverse). Values denote F1 scores on the test split.}
  \label{fig:rl_reverse_original}
\end{figure*}

We provide additional results on how shift vectors encode task outcomes. Table~\ref{tab:shift-probe-all} shows that even a single sublayer--timestep direction can be highly predictive, reaching ROC-AUCs of $96.9$--$99.0$ on LIBERO-Spatial, LIBERO-Object, and ManiSkill. Across individual positions, some directions show larger projections for successful episodes, while others show the opposite. These two cases appear in similar proportions for both base and RL policies (Table~\ref{tab:reverse_original}). Figures~\ref{fig:base_reverse_original} and~\ref{fig:rl_reverse_original} visualize this structure across sublayer--timestep positions. This suggests that RL may modulate task-relevant representations by increasing or decreasing activation along different shift directions across sublayers and timesteps, with their combined effects shaping successful policy behavior.

\begin{table}[htbp!]
    \centering
    \small
    \setlength{\tabcolsep}{4.5pt}
    \renewcommand{\arraystretch}{0.95}
    \caption{\textbf{Extended linear probing results with shift vectors in \boldmath$\pi_{0.5}$.}
    F1 and ROC-AUC (\%) for probes applied to the base and RL policies. We use per-benchmark 70:30 train--test splits.
    Logistic denotes our $\ell_2$-regularized logistic regression probes. Single-Position uses a single sublayer--timestep shift vector, with the position selected by the best F1 on the training split. Random denotes probes trained on randomly shuffled outcome labels.}
    \label{tab:shift-probe-all}

    \begin{tabular}{@{}llcccc@{}}
        \toprule
        & &
        \multicolumn{2}{c}{\textbf{Base Policy}} &
        \multicolumn{2}{c}{\textbf{RL Policy}} \\
        \cmidrule(lr){3-4} \cmidrule(lr){5-6}
        Benchmark & Method
        & F1 & AUC
        & F1 & AUC \\
        \midrule

        \multirow{3}{*}{LIBERO Spatial}
        & Logistic
        & \textbf{97.1} & \textbf{99.6}
        & \textbf{96.3} & \textbf{96.6} \\
        & Single-Position
        & 93.6 & 99.0
        & 85.7 & 78.0 \\
        & Random
        & $58.0 \pm 20.5$ & $46.5 \pm 4.0$
        & $75.2 \pm 13.6$ & $51.5 \pm 3.9$ \\
        \midrule

        \multirow{3}{*}{LIBERO Object}
        & Logistic
        & \textbf{98.2} & \textbf{98.6}
        & \textbf{97.3} & \textbf{97.4} \\
        & Single-Position
        & 97.8 & 98.2
        & \textbf{97.3} & 94.7 \\
        & Random
        & $76.3 \pm 10.2$ & $53.4 \pm 6.3$
        & $86.2 \pm 15.8$ & $47.2 \pm 13.3$ \\
        \midrule

        \multirow{3}{*}{LIBERO Goal}
        & Logistic
        & \textbf{77.5} & \textbf{98.9}
        & \textbf{74.9} & \textbf{68.0} \\
        & Single-Position
        & 70.5 & 51.2
        & 72.5 & 60.0 \\
        & Random
        & $47.6 \pm 37.2$ & $49.0 \pm 3.5$
        & $65.4 \pm 28.9$ & $52.6 \pm 6.4$ \\
        \midrule

        \multirow{3}{*}{ManiSkill}
        & Logistic
        & \textbf{95.8} & \textbf{99.6}
        & \textbf{98.3} & \textbf{98.3} \\
        & Single-Position
        & 90.7 & 96.9
        & 95.6 & 92.0 \\
        & Random
        & $42.9 \pm 8.9$ & $51.8 \pm 7.0$
        & $87.1 \pm 3.4$ & $46.5 \pm 9.3$ \\
        \midrule

        \multirow{3}{*}{MetaWorld}
        & Logistic
        & \textbf{58.3} & \textbf{74.2}
        & 81.1 & \textbf{90.1} \\
        & Single-Position
        & 53.1 & 50.6
        & \textbf{82.4} & 72.0 \\
        & Random
        & $43.4 \pm 7.8$ & $49.4 \pm 4.8$
        & $64.8 \pm 6.6$ & $49.4 \pm 6.1$ \\

        \bottomrule
    \end{tabular}
\end{table}

\begin{table}[!htb]
    \centering
    \small
    \setlength{\tabcolsep}{5pt}
    \renewcommand{\arraystretch}{0.95}
    \caption{\textbf{Probe direction at single positions in \boldmath$\pi_{0.5}$.}
Fraction of single positions where the original or reverse probe direction is predictive (\%) across benchmarks and policies. Original indicates that larger projections are associated with success, whereas Reverse indicates the opposite.}
    \label{tab:reverse_original}
    \begin{tabular}{llcc}
        \toprule
        Model & Benchmark & Original (+) & Reverse ($-$) \\
        \midrule
        \multirow{6}{*}{Base}
        & LIBERO Spatial & 58.4 & 41.6 \\
        & LIBERO Object  & 62.7 & 37.3 \\
        & LIBERO Goal    & 43.2 & 56.8 \\
        & ManiSkill      & 44.6 & 55.4 \\
        & MetaWorld      & 52.4 & 47.6 \\
        & \textbf{Average} & \textbf{52.6} & \textbf{47.4} \\
        \midrule
        \multirow{6}{*}{RL}
        & LIBERO Spatial & 58.9 & 41.1 \\
        & LIBERO Object  & 55.1 & 44.9 \\
        & LIBERO Goal    & 50.8 & 49.2 \\
        & ManiSkill      & 36.5 & 63.5 \\
        & MetaWorld      & 37.8 & 62.2 \\
        & \textbf{Average} & \textbf{48.3} & \textbf{51.7} \\
        \bottomrule
    \end{tabular}
\end{table}

\clearpage

\begin{figure*}[!htb]
  \centering
  \includegraphics[width=0.8\linewidth]
  {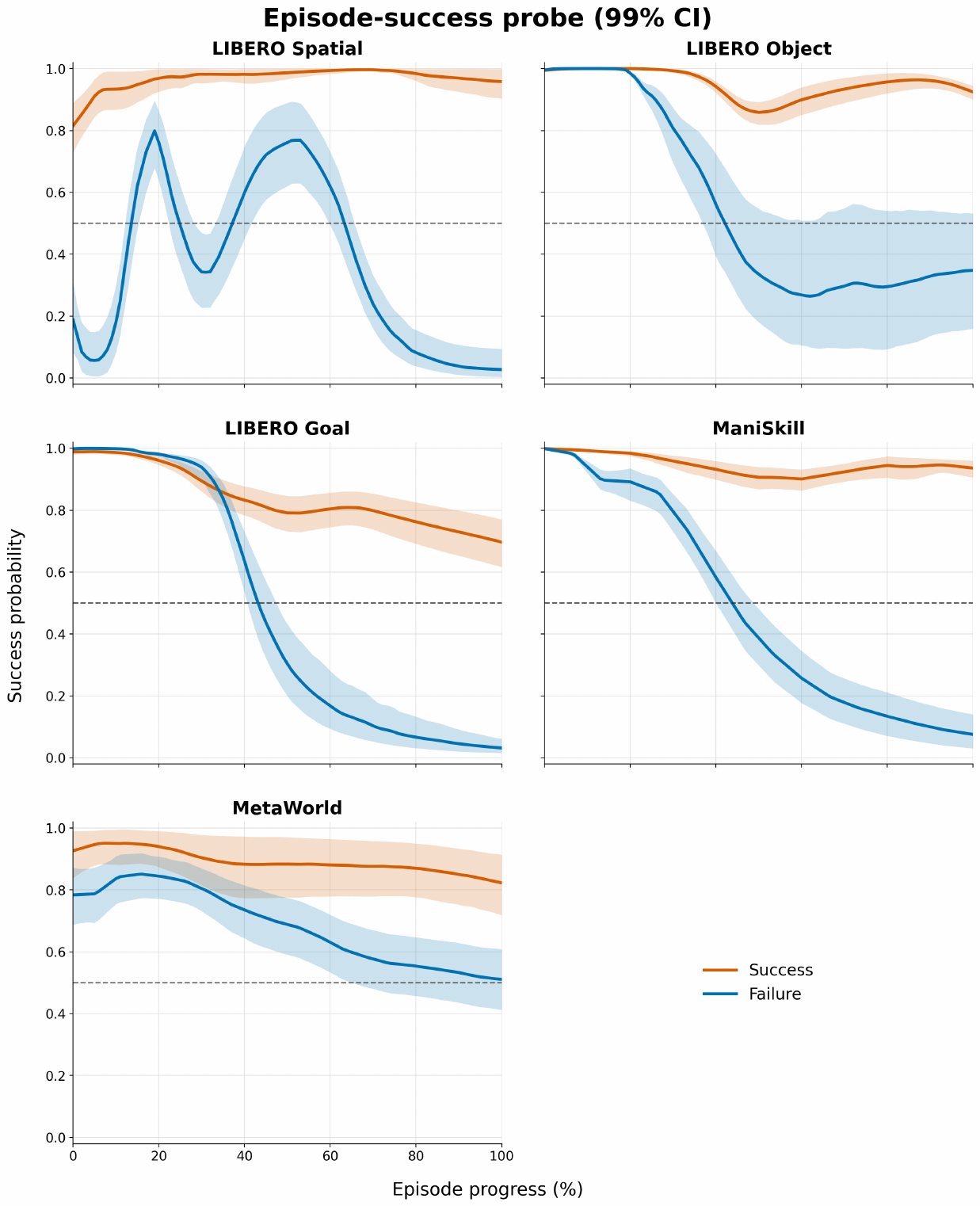}
  \caption{\textbf{Episode-level probe trajectories across different \boldmath$\pi_{0.5}$ checkpoints.}
Episode progress is normalized to $0$--$100\%$ for each action horizon. Probe outputs from successful (orange) and failed (blue) episodes are interpolated over normalized progress, and we plot the mean prediction with $95\%$ confidence intervals.}
  \label{fig:probe-trajectory}
\end{figure*}

\clearpage
We further show the episode-level trajectories of our logistic regression probe in Figure~\ref{fig:probe-trajectory}. These results show that task-outcome signals arise before the end of an episode, as predicted success begins to separate between successful and failed trajectories early in execution.

For the logistic regression probes, we sweep the $\ell_2$-regularization hyperparameter over $\{0.0001, 0.0003, 0.001, 0.003, 0.01, 0.03, 0.1, 0.3, 1, 3, 10, 30, 100\}$ and select the best value via cross-validation on the training split.

\clearpage

\subsection{Steering Results}
\label{app:steering_results}
We examine whether steering along the shift directions can directly influence policy behavior. Steering the base policy along the learned shift directions produces substantial gains, supporting their causal relevance (Table~\ref{tab:base-steering-sweep}). Here, steering is applied simultaneously to all sublayer--timestep pairs. Although the optimal coefficient $\alpha$ varies across benchmarks, performance generally improves as $\alpha$ approaches the optimum and declines once steering becomes too strong. These results show that shift vectors recover a substantial portion of the RL improvement when applied to the base policy.

\begin{table}[!htb]
    \centering
    \small
    \setlength{\tabcolsep}{4.5pt}
    \renewcommand{\arraystretch}{0.95}
\caption{\textbf{Base-policy steering across different strengths in \boldmath$\pi_{0.5}$.}
Success rates (\%) across steering coefficients $\alpha$, where $\alpha$ scales the RL-induced shift vector before addition. Values in parentheses denote changes relative to the base policy. Best Recovery is the fraction of the RL improvement recovered by the best steering result.}
    \label{tab:base-steering-sweep}
    
    \begin{tabular}{@{}lccccccc@{}}
        \toprule
        Benchmark
        & Base
        & RL
        & \multicolumn{4}{c}{Steering coefficient $\alpha$}
        & Best Recovery \\
        \cmidrule(lr){4-7}
        & & &
        0.5 & 1.0 & 1.5 & 2.0
        & (\%) \\
        \midrule

        LIBERO Goal
        & 82.8
        & 93.6
        & \textbf{88.2 \gain{5.4}}
        & 84.4 \gain{1.6}
        & 79.8 \drop{3.0}
        & 73.2 \drop{9.6}
        & 50.0 \\

        LIBERO Object
        & 95.4
        & 99.2
        & 95.2 \drop{0.2}
        & 97.6 \gain{2.2}
        & 98.0 \gain{2.6}
        & \textbf{98.8 \gain{3.4}}
        & 89.5 \\

        LIBERO Spatial
        & 85.0
        & 96.6
        & 86.4 \gain{1.4}
        & \textbf{92.0 \gain{7.0}}
        & 90.0 \gain{5.0}
        & 90.2 \gain{5.2}
        & 60.3 \\

        ManiSkill
        & 42.2
        & 89.1
        & 48.4 \gain{6.2}
        & 55.3 \gain{13.1}
        & 57.5 \gain{15.3}
        & \textbf{63.8 \gain{21.6}}
        & 46.1 \\

        MetaWorld
        & 42.8
        & 69.2
        & 56.4 \gain{13.6}
        & \textbf{59.4 \gain{16.6}}
        & 58.6 \gain{15.8}
        & 57.2 \gain{14.4}
        & 62.9 \\

        \bottomrule
    \end{tabular}
\end{table}

\end{document}

%% file: math_commands.tex
\usepackage{amsmath,amsfonts,bm}

\def\eqref#1{equation~\ref{#1}}
\def\1{\bm{1}}

\DeclareMathAlphabet{\mathsfit}{\encodingdefault}{\sfdefault}{m}{sl}
\SetMathAlphabet{\mathsfit}{bold}{\encodingdefault}{\sfdefault}{bx}{n}